\documentclass{article} % For LaTeX2e
\usepackage{arxiv,times}[]

\usepackage{amsmath,amsfonts,bm}

\def\eqref#1{equation~\ref{#1}}
\def\1{\bm{1}}

\DeclareMathAlphabet{\mathsfit}{\encodingdefault}{\sfdefault}{m}{sl}
\SetMathAlphabet{\mathsfit}{bold}{\encodingdefault}{\sfdefault}{bx}{n}

\newcommand{\softmax}{\mathrm{softmax}}

\DeclareMathOperator*{\argmax}{arg\,max}

\usepackage{hyperref}
\usepackage{url}
\usepackage{algorithm}
\usepackage{algorithmic}
\usepackage{multirow} % for professional tables
\usepackage{booktabs} % for professional tables
\usepackage{colortbl}
\usepackage{xcolor} % 用于颜色定义
\usepackage{amsmath} % 用于数学符号
\usepackage{amssymb} % 用于黑色三角符号
\usepackage{amsthm} % for proof

\newtheorem{proposition}{Proposition}

\usepackage{bbold}
\usepackage{subcaption}
\usepackage{bbm}
\usepackage{graphicx}
\usepackage{subcaption}  % For subfigures
\usepackage{wrapfig}
\usepackage{graphicx} % Needed to include images

\definecolor{green}{rgb}{0.0, 0.5, 0.0}
\definecolor{red}{rgb}{0.5, 0.0, 0.0}
\definecolor{lightgray}{gray}{0.9}

\newcommand{\cellgray}{\cellcolor{lightgray}}

\title{Consistency Calibration: \\Improving Uncertainty Calibration via Consistency among Perturbed Neighbors}

\author{Linwei~Tao\\
University of Sydney\\
\texttt{linwei.tao@sydney.edu.au} \\
\And
Haolan~Guo\\
University of Sydney\\
\texttt{hguo4658@uni.sydney.edu.au} \\
\And
Minjing Dong \\
City University of Hong Kong \\
\texttt{minjdong@cityu.edu.hk}
\And
Chang Xu \\
University of Sydney\\
\texttt{c.xu@sydney.edu.au}
}

\iclrfinalcopy % Uncomment for camera-ready version, but NOT for submission.
\begin{document}

\maketitle

\begin{abstract}
Calibration is crucial in deep learning applications, especially in fields like healthcare and autonomous driving, where accurate confidence estimates are vital for decision-making. However, deep neural networks often suffer from miscalibration, with reliability diagrams and Expected Calibration Error (ECE) being the only standard perspective for evaluating calibration performance. In this paper, we introduce the concept of \textit{consistency} as an alternative perspective on model calibration, inspired by uncertainty estimation literature in large language models (LLMs). We highlight its advantages over the traditional reliability-based view. Building on this concept, we propose a post-hoc calibration method called Consistency Calibration (CC), which adjusts confidence based on the model's consistency across perturbed inputs. CC is particularly effective in locally uncertainty estimation, as it requires no additional data samples or label information, instead generating input perturbations directly from the source data. Moreover, we show that performing perturbations at the logit level significantly improves computational efficiency. We validate the effectiveness of CC through extensive comparisons with various post-hoc and training-time calibration methods, demonstrating state-of-the-art performance on standard datasets such as CIFAR-10, CIFAR-100, and ImageNet, as well as on long-tailed datasets like ImageNet-LT. 
% Code is available at \hyperlink{here}{https://anonymous.4open.science/r/Consistency-Calibration-E248}.
\end{abstract}
\section{Introduction}
Calibration is essential in many deep learning applications where accurate confidence estimates are as important as the predictions themselves. In fields like healthcare~\cite{chen2018calibration} and autonomous driving~\cite{feng2019can}, decisions often rely not only on the model's output but also on how confident the model is in its predictions. A well-calibrated model should reflect the ground truth uncertainty. In healthcare, for instance, a model that accurately reflects uncertainty can help doctors trust the system’s confidence when diagnosing critical conditions. 

However, current deep learning models are often found to be miscalibrated~\citep{guo2017calibration}. To evaluate calibration performance, \citet{naeini2015obtaining} introduced ECE, which has become the gold standard, based on the reliability diagram~\citep{degroot1983comparison}. Although several improved metrics have since been proposed, such as AdaptiveECE (AdaECE)~\citep{nixon2019measuring} and ClasswiseECE (CECE)~\citep{kull2019beyond}, they all adopt the same fundamental perspective on calibration: if a model assigns 80\% confidence to its predictions, then, ideally, 80\% of those predictions should be correct. We refer to this classical approach as the \textit{reliability view}, which seeks to align predicted confidence levels with actual model accuracy.

The concept of \textit{consistency} has gained increasing importance in black-box uncertainty estimation, particularly in recent developments in large language models (LLMs)~\citep{wang2022self, tam2022evaluating, xiong2023can, geng2023survey}. If an LLM is confident in its answer, it should provide consistent responses to similar questions. For instance, if an LLM confidently answers the question ``What is the answer to 5 + 3?" with ``8", it should also consistently provide ``8" for the similar question ``What is the result of five plus three?" In this paper, we extend this concept of consistency to model calibration, proposing a new perspective of calibration called consistency.

Specifically, in a classification task, if a model is confident in its prediction, it should consistently provide the same output across multiple perturbed versions of the input. Consistency measures how often a model's prediction remains unchanged when the input is perturbed within a small neighborhood. A high consistency score implies that the model's predictions are stable and confident. In this view, a perfectly calibrated model should have its predicted confidence levels align with the consistency observed across these perturbed inputs.

In the following sections, we discuss the differences between calibration from the perspectives of reliability and consistency in Sections~\ref{Calibration in the View of Reliability} and~\ref{Calibration in the view of Consistency}. Section~\ref{Consistency as a Representation of Ground Truth Uncertainty} highlights the advantages of the consistency approach over the reliability view through a toy example. In Section~\ref{More Efficient Consistency Calibration}, CC is introduced, which involves perturbing the logits. We provide empirical evidence to explain its effectiveness in Section~\ref{Why consistency calibration works?}. Finally, in Section~\ref{Consistency as a Local Uncertainty Estimation}, we demonstrate that consistency can serve as a reliable method for local uncertainty estimation.

Our contributions can be summarized as follows:

\begin{itemize}
    \item We introduce a novel perspective on calibration based on consistency and highlight its advantages over traditional reliability view represented by ECE.
    \item We propose an easy-to-implement and computationally efficient post-hoc calibration method called \textit{Consistency Calibration}, which replaces the original confidence score with a consistency measure calculated from perturbed logits using data neighbors.
    \item \textit{CC} serves as a reliable and effective method for local uncertainty estimation, as it does not require additional data samples or label information. Instead, it generates data neighborhoods based on the source data.
    \item We conduct comparisons with multiple post-hoc and training-time calibration methods, demonstrating state-of-the-art performance on standard datasets, including CIFAR-10, CIFAR-100, and ImageNet, as well as in long-tailed scenarios like ImageNet-LT.
\end{itemize}

\section{Methodology}
In a classification task, let \( \mathcal{X} \) represent the input space and \( \mathcal{Y} \) the label space. The neural network \(f(\cdot)\) and projection head \(g(\cdot)\) maps \( x \in \mathcal{X} \) to a vector of logits \( z=g(f(x)) \in \mathbb{R}^K \), where each \( z_k \) is the logit for class \( k \). These logits are then transformed into a probability distribution \( \hat{p} = \text{softmax}(z) \) over \( K \) classes using the softmax function:
\begin{equation}
\hat{p}_k = \frac{e^{z_k}}{\sum_{i=1}^{K} e^{z_i}}, \quad k = 1, \dots, K,    
\end{equation}
where \( \mathbf{k} = \arg\max_i \hat{p}_i \) denotes the predicted label index. The ground-truth label \( y \in \mathcal{Y} \) represents the true class, and \( \hat{y} \in \mathcal{Y} \) is the predicted label. The confidence score \( \hat{p}_\mathbf{k} \) represents the predicted probability assigned to the predicted label \( \mathbf{k} \). 
% For simplicity, we omit the subscript \( \mathbf{k} \) in the following content.

\subsection{Calibration in the View of Reliability}  
\label{Calibration in the View of Reliability}
Calibration in the view of reliability has been widely accepted since the introduction of the reliability diagram by \citet{degroot1983comparison}. In this view, a classifier is considered perfectly calibrated if its predicted confidence \( \hat{p} \) accurately represents the true probability of correctness. Formally, this is expressed as:
\begin{equation}
\mathbb{P}(\hat{y} = y \mid \hat{p} = p) = p \quad \text{for all} \, p \in [0, 1].
\end{equation}
In other words, if a model assigns a confidence score of 80\%, the prediction \( \hat{y} \) should be correct 80\% of the time.
To move beyond visual inspection of reliability diagram, \citet{naeini2015obtaining} developed a quantitative metric from the reliability diagram called the \textit{Expected Calibration Error} (ECE). ECE provides a more precise measurement of miscalibration by calculating the average discrepancy between a model's predicted confidence and the actual accuracy of predictions at the same confidence level. ECE is defined as:
\begin{equation}
\text{ECE} = \mathbb{E}_{\hat{p}} \left[ \left| \mathbb{P}(\hat{y} = y \mid \hat{p}) - \hat{p} \right| \right].
\end{equation}
In practice, due to finite sample sizes, an approximation is used by binning predictions into \( M \) equally spaced confidence intervals, \( \{B_m\}_{m=1}^M \). Each bin \( B_m \) contains predictions with confidence scores \( \hat{p} \in \left[ \frac{m}{M}, \frac{m+1}{M} \right) \). For each bin, the average confidence \( C_m \) and accuracy \( A_m \) are computed as:
\begin{equation}
    C_m = \frac{1}{|B_m|} \sum_{i \in B_m} \hat{p}_i,\quad A_m = \frac{1}{|B_m|} \sum_{i \in B_m} \mathbbm{1}(\hat{y}_i = y_i),
    \label{eq:ece_conf_accuracy}
\end{equation}
where \( \mathbbm{1} \) is the indicator function, and \( |B_m| \) is the number of samples in bin \( B_m \).
The approximate ECE is then computed as the weighted average of the absolute difference between bin accuracy and bin confidence:
\begin{equation}
\text{ECE} = \sum_{m=1}^M \frac{|B_m|}{N} \left| A_m - C_m \right|,
\end{equation}
where \( N \) is the total number of samples. 
Several variants of ECE exist. For instance, \textit{AdaECE} uses adaptive binning to ensure equal sample sizes in each bin and avoid the issue of uneven confidence distribution in ECE, while \textit{CECE} computes ECE on a per-class basis, enabling better detection of class-specific calibration errors.

\subsection{Calibration in the view of Consistency}
\label{Calibration in the view of Consistency}
We offer an alternative perspective on calibration by examining it through the concept of consistency. In a real-world scenario, an individual confident in their answer tends to maintain that answer, even when faced with external doubts or minor alterations to the question. On the other hand, someone who is uncertain might change their response when presented with slightly misleading information or variations in the question. We define this adherence to the original answer as \textit{consistency}. 

Recent advances in LLMs, particularly black-box models utilize \textit{factual consistency} to enhance performance \citep{wang2022self, tam2022evaluating, xiong2023can, geng2023survey}. These studies frame the consistency of a model's responses as an indicator of its uncertainty. In the context of classification tasks, calibration can also be described in terms of consistency. Specifically, for classification models, we can formalize this relationship as follows:
\begin{proposition}
\label{prop:definition of well calibration in the view of consistency}
If a model is confident in its prediction, it should consistently output the same prediction when the input is slightly perturbed. The consistency \( c \) of a sample \( x \) is defined as
\begin{equation}
     c_k(x) = \frac{1}{T} \sum_{t=1}^{T} \mathbbm{1}(\hat{y}(\Tilde{x}_t) = k), \text{ where } d(\Tilde{x}_t, x) < \epsilon^*,  \quad \text{for} \, k = 1, \dots, K
\end{equation}
where \( T \) is the number of perturbed neighbors, \( \hat{y}(\Tilde{x}_t) \) is the predicted label for the perturbed input \( \Tilde{x}_t \), and the distance between the original sample \( x \) and its perturbed version \( \Tilde{x}_t \) is smaller than a constant \( \epsilon^* \), according to some distance metric \( d \).
A model is said to be perfectly calibrated if, for all samples \( x \), given a suitable set of perturbed neighbors \( \{\Tilde{x}_t \mid t = 1, \dots, T\} \), the predicted confidence score \( \hat{p}(x) \) satisfies:
\begin{equation}
    \hat{p}_k(x) = c_k(x), \quad \text{for} \, k = 1, \dots, K
\end{equation}
\end{proposition}

However, identifying a suitable perturbed neighborhood is non-trivial—it is challenging to determine an appropriate constant \( \epsilon^* \) and distance metric \( d \). Fortunately, in image classification tasks, a perturbed neighbor is often considered a data-augmented version of the original image. Thus, we begin our exploration by using image data augmentation.

To evaluate the effectiveness of consistency-based confidence, we design an experimental setting using a ResNet-50 model trained on CIFAR-10 with data augmentation (RandomCrop and RandomHorizontalFlip). We generate perturbed neighbors by applying various levels of data augmentation to the entire CIFAR-10 test set, creating 100 perturbed neighbors for each test sample. The calibration performance of consistency is assessed on the test set in the following settings:
\begin{itemize}
    \item \textit{Baseline}: Confidence score is extracted on the original test set, serving as the baseline.
    \item \textit{Weak Augmentation (Train Augmentation)}: The confidence score is replaced with consistency derived from perturbed neighbors generated using train-time augmentation (RandomCrop and RandomHorizontalFlip), denoted by the yellow star.
    \item \textit{Moderate Augmentation (Train Augmentation + ColorJitter)}: The confidence score is replaced with consistency measured from perturbed neighbors generated using train-time augmentation and varying strengths of ColorJitter, as indicated by the x-axis values.
    \item \textit{Stronger Augmentation (Train Augmentation + ColorJitter + Blur)}: The confidence score is replaced with consistency measured from perturbed neighbors generated using train-time augmentation, ColorJitter, and Blur, represented by the red triangle.
\end{itemize}

\begin{figure}[ht]
    \centering
    \begin{subfigure}[b]{0.35\textwidth}
        \centering
        \includegraphics[width=\textwidth]{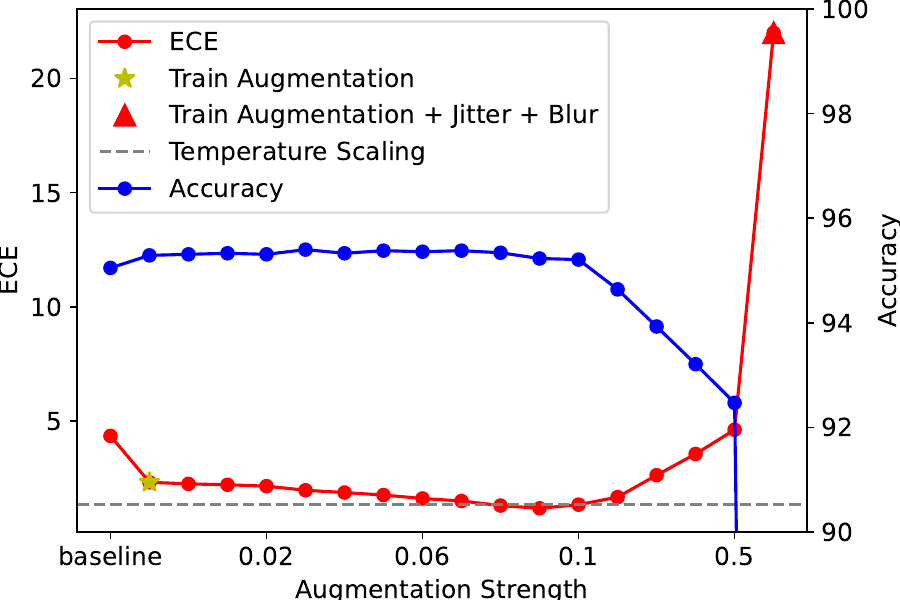}
        \caption{Perturbation applied to images with different augmentations. Consistency calculated using weak or moderate augmentation neighbors significantly reduces calibration error. } 
        \label{fig:image_disturb}
    \end{subfigure}
    \hfill
    \begin{subfigure}[b]{0.29\textwidth}
        \centering
        \includegraphics[width=\textwidth]{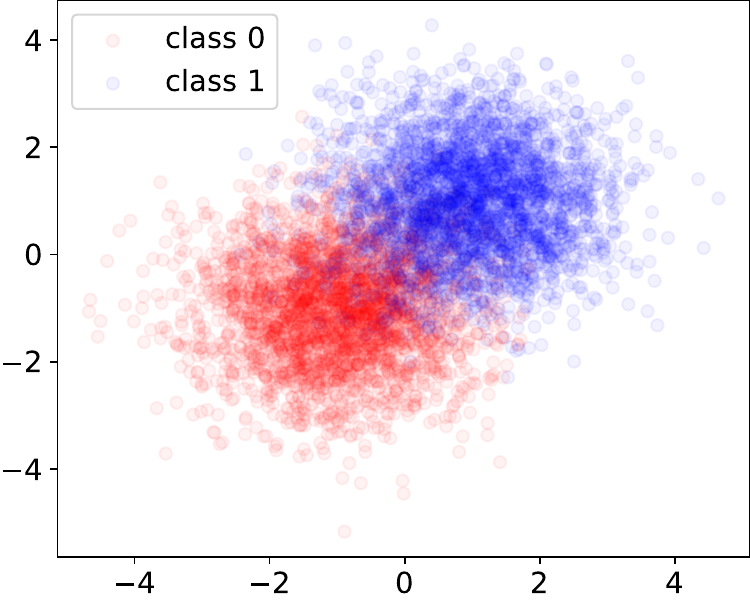}
        \caption{Toy dataset generated from two two-dimensional Gaussian distributions. Samples near the diagonal are uncertain to belong to class 0 or class 1.} 
        \label{fig:data_distribution}
    \end{subfigure}
    \hfill
    \begin{subfigure}[b]{0.31\textwidth}
        \centering
        \includegraphics[width=\textwidth]{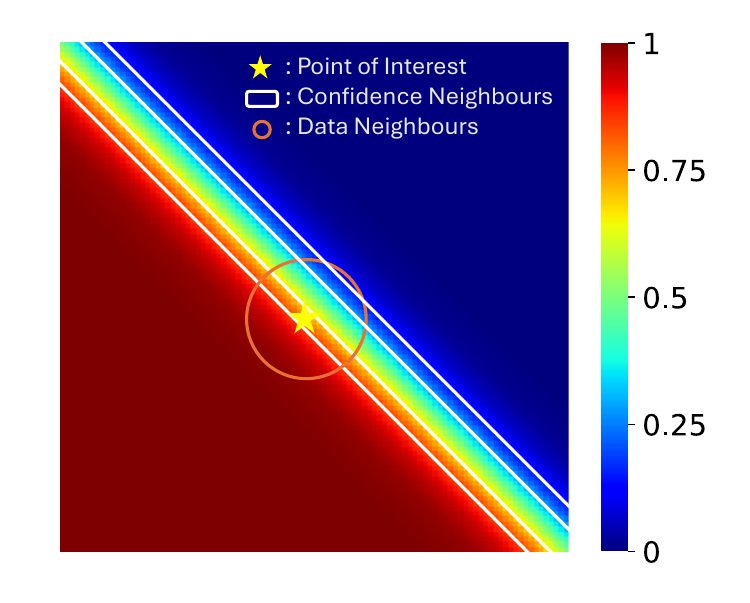}
        \caption{Heatmap of ground truth uncertainty calculated from the PDF, as given by Eq.~\ref{eq:ground_truth_uncertainty}. Circle and box indicate different neighbourhood selection criteria.} 
        \label{fig:ground_truth_confidence}
    \end{subfigure}
    \label{fig:input_disturb}
    \caption{Illustrations of Consistency, Toy Data Distributions, and Ground Truth Uncertainty.}
    \vspace{-0.2in}
\end{figure}

The evaluation results are shown in Figure~\ref{fig:image_disturb}. Consistency using neighbors generated with weak augmentation significantly reduces calibration error compared to the baseline. As we increase the perturbation strength with moderate augmentation, as shown by the x-axis values, the calibration error continues to decrease with minimal impact on accuracy, outperforming the commonly used calibration method, Temperature Scaling, up to a certain perturbation threshold.

However, when moderate augmentation with strength exceeds 0.1, accuracy begins to decline, and ECE increases sharply. With neighbors generated from Stronger Augmentation, both calibration and prediction accuracy deteriorate. This likely occurs because stronger perturbations distort the input to the extent that the model can no longer recognize the data, leading to degraded performance. This suggests that consistency has the potential to provide accurate uncertainty estimates when a suitable perturbed neighborhood is identified.

\subsection{Consistency as a Representation of Ground Truth Uncertainty} 
\label{Consistency as a Representation of Ground Truth Uncertainty}
On one hand, the \textit{reliability approach} estimates calibration error by comparing the prediction confidence with the average correctness of samples that have similar confidence levels. In this view, the average correctness of such sample neighborhood is treated as an approximation of the ground truth uncertainty. On the other hand, the \textit{consistency approach} directly uses consistency as a measure of ground truth uncertainty. Thus, we are interested in determining which of these two approaches more accurately approximates this uncertainty.

To explore this, we constructed a toy dataset consisting of two two-dimensional Gaussian distributions representing two groups of data:
\(
\mathcal{N}(\mu_0, \Sigma), \mathcal{N}(\mu_1, \Sigma)
\)
where \( \mu_0 \) and \( \mu_1 \) are the mean vectors, and \( \Sigma \) is the shared covariance matrix for both groups, labeled 0 and 1, respectively. We generated 1,000,000 data points from each group to form the training dataset, which was used to train a CNN model. An additional 50,000 samples from each group were used to create the test dataset. The input space is \( \mathcal{X} = \mathbb{R}^2 \), and the label space is \( \mathcal{Y} = \{0, 1\} \), as illustrated in Figure~\ref{fig:data_distribution}.

The ground truth uncertainty, \( \eta(x) \), is calculated from the probability density function (PDF) of each distribution:
\begin{equation}
\eta(x) = \frac{p^0(x)}{p^0(x) + p^1(x)}
\label{eq:ground_truth_uncertainty}
\end{equation}
where \( p^0(\cdot) \) and \( p^1(\cdot) \) are the PDFs of the two distributions. The ground truth uncertainty is illustrated in Figure~\ref{fig:ground_truth_confidence}. For each label, the ground truth confidence can be expressed as \( (\eta(x), 1 - \eta(x)) \).

In Figure~\ref{fig:ground_truth_confidence}, for a point of interest (marked by a star), the reliability-based approach estimates ground truth uncertainty by calculating the average correctness \( A = \frac{1}{|B|} \sum_{i \in B} \mathbbm{1}(\hat{y}_i = y_i), \) over a ``confidence neighborhood" \( B \) (i.e., samples with similar confidence, enclosed by the white boxes), similar to the definition Eq.~\ref{eq:ece_conf_accuracy} in ECE. In contrast, the consistency approach estimates uncertainty by considering ``data neighborhood," as illustrated by the orange circle. While the reliability approach relies on the availability of multiple data samples within the confidence neighborhood, the consistency approach generates data neighborhoods by perturbing the data.

\begin{figure}[ht]
    \centering
    \begin{subfigure}[b]{0.315\textwidth}
        \centering
        \includegraphics[width=\textwidth]{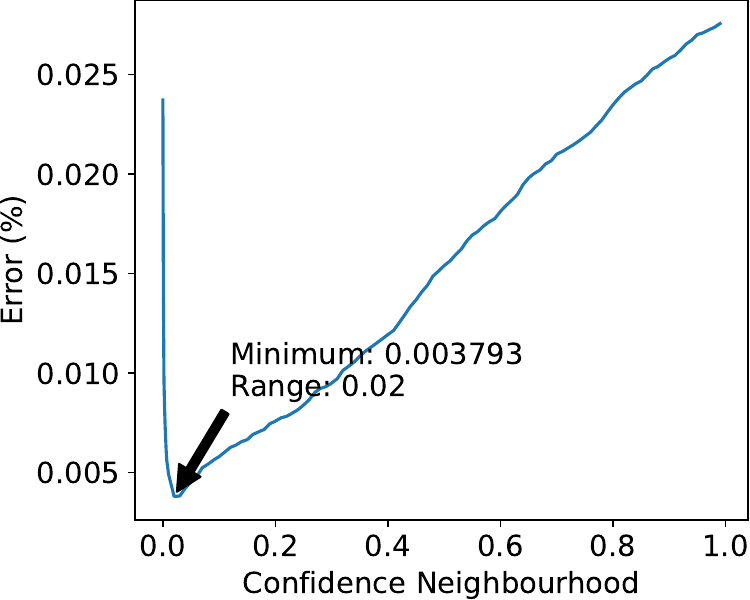}
        \caption{\textbf{Reliability view:} Estimating ground truth uncertainty using neighbors with confidence differences indicated on the x-axis.}
        \label{fig:confidence_neighbour_error}
    \end{subfigure}
    \hfill
    \begin{subfigure}[b]{0.315\textwidth}
        \centering
        \includegraphics[width=\textwidth]{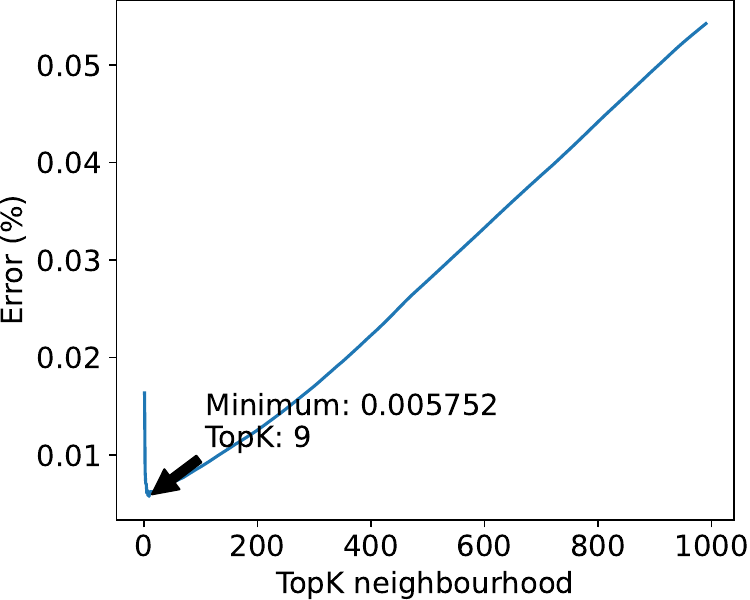}
        \caption{\textbf{Reliability view:} Estimating ground truth uncertainty using the top-k nearest confidence neighbors as indicated on the x-axis.}
        \label{fig:topk_neighbour_error}
    \end{subfigure}
    \hfill
    \begin{subfigure}[b]{0.315\textwidth}
        \centering
        \includegraphics[width=\textwidth]{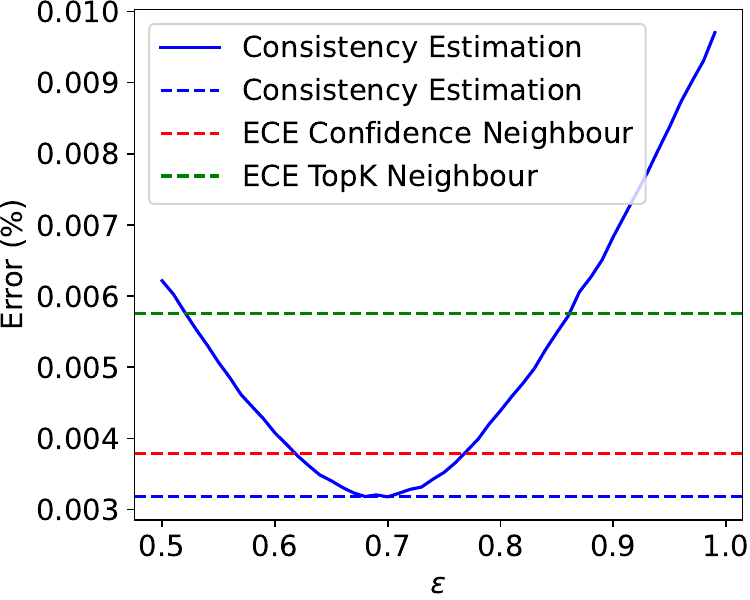}
        \caption{\textbf{Consistency view:} Estimating ground truth uncertainty using data neighbors perturbed within \( \epsilon \) as indicated on the x-axis.}
        \label{fig:consistency_ece_comparison}
    \end{subfigure}
    \caption{Comparison of Consistency vs. Reliability in Estimating Ground Truth Uncertainty}
    \label{fig:estimate_gt_uncertainty}
    \vspace{-0.2in}
\end{figure}

The key differences between the reliability and consistency views lie in their neighborhood selection criteria \( S \) and aggregation methods. The reliability view selects a neighborhood \( B \) based on confidence similarity and aggregates the correctness of the samples, while the consistency view selects a neighborhood based on data perturbations and computes consistency, as described in Eq.~\ref{eq:consistency_calibration}. To compare the two approaches, we evaluate them under three neighborhood selection criteria:

\begin{itemize}
    \item \textit{Figure~\ref{fig:confidence_neighbour_error}:} Reliability view (ECE): \( B = \{\tilde{x} \mid |\hat{p}(x) - \hat{p}(\tilde{x})| < \epsilon \} \)
    \item \textit{Figure~\ref{fig:topk_neighbour_error}:} Reliability view (AdaECE): \( B = \{\tilde{x} \mid \text{Top K closest confidence neighbors} \} \)
    \item \textit{Figure~\ref{fig:consistency_ece_comparison}:} Consistency view (perturbing data): \( B = \{\tilde{x}  \mid \tilde{x} = x+\epsilon \} \)
\end{itemize}

In Figures~\ref{fig:confidence_neighbour_error} and \ref{fig:topk_neighbour_error}, we use the reliability approach to approximate ground truth uncertainty based on two confidence-neighbor selection criteria. In Figure~\ref{fig:confidence_neighbour_error}, we replicate the standard ECE~\citep{guo2017calibration} approach by selecting confidence neighbors solely based on confidence differences. The x-axis represents the allowed confidence difference between neighbors and the point of interest, while the y-axis shows the average error between estimated and ground truth uncertainty across the test set. In Figure~\ref{fig:topk_neighbour_error}, we replicate the AdaECE~\citep{nixon2019measuring} approach by selecting the top-K nearest confidence neighbors to estimate uncertainty, with the lowest error (0.57\%) achieved by selecting the top 9 nearest neighbors.

In Figure~\ref{fig:consistency_ece_comparison}, we apply Gaussian noise \( \epsilon \) to perturb the data samples and compute consistency across 100 generated neighbors, with the x-axis representing the noise strength. We compare the uncertainty estimates from the consistency approach with those from the reliability approach. The dashed lines indicate the minimal error achieved by each method. Within a certain range of perturbation strengths, the consistency approach outperforms, yielding a ground truth uncertainty estimation with an overall error as low as 0.3\%.

It is important to note that the reliability approach using confidence neighborhoods is essentially equivalent to the ECE measurement, where the allowed confidence gap functions similarly to the hyperparameter ``number of bins" in ECE. As shown in Figure~\ref{fig:estimate_gt_uncertainty}, the estimation error is sensitive to the allowed confidence gap—meaning that the choice of ``number of bins" can significantly impact the ability of ECE to estimate the ground truth uncertainty. Similarly, this sensitivity is also observed in the consistency method, where the strength of perturbation noise affects the uncertainty approximation. Despite this sensitivity, the consistency approach achieves a lower overall estimation error, suggesting its potential as a robust alternative calibration metric.

\subsection{More Efficient Consistency Calibration}
\label{More Efficient Consistency Calibration}
\begin{figure}[ht]
    \centering
    \begin{subfigure}[b]{0.36\textwidth}
        \centering
        \includegraphics[width=\textwidth]{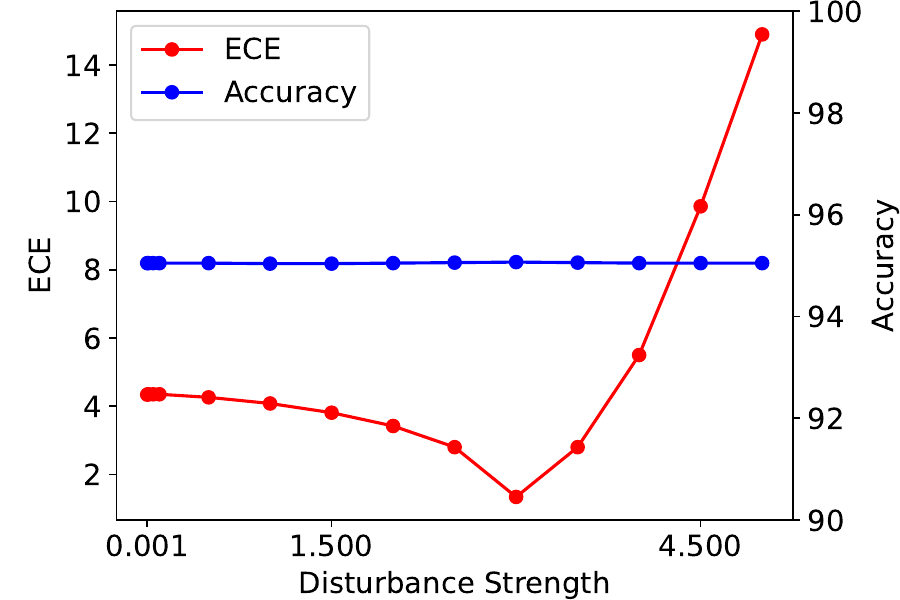}
        \caption{Performance of consistency calibration using data neighbors with feature perturbations at varying noise levels.}
        \label{fig:feature_disturb}
    \end{subfigure}
    \hfill
    \begin{subfigure}[b]{0.36\textwidth}
        \centering
        \includegraphics[width=\textwidth]{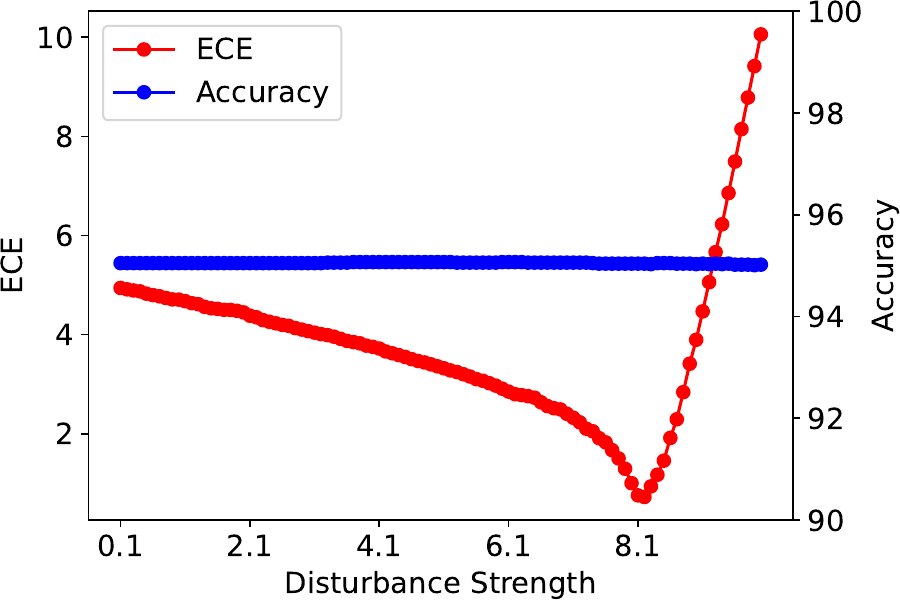}
        \caption{Performance of consistency calibration using data neighbors with logit perturbations at varying noise levels.}
        \label{fig:logit_disturb}
    \end{subfigure}
    \hfill
    \begin{subfigure}[b]{0.25\textwidth}
        \centering
        \includegraphics[width=\textwidth]{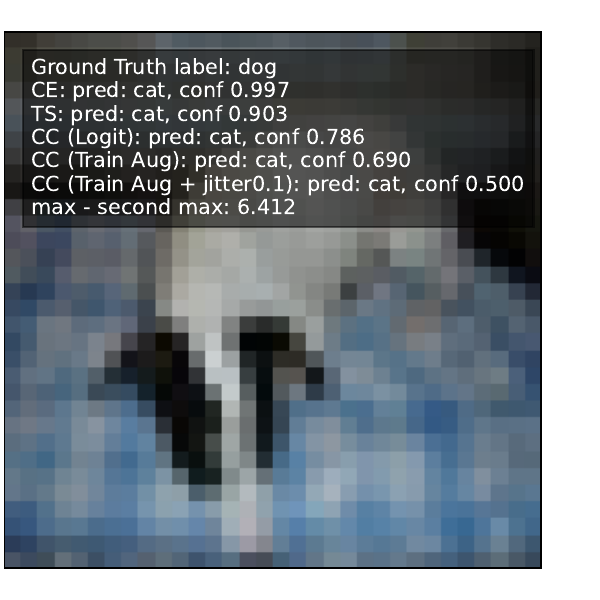}
        \caption{Performance of local uncertainty estimation using consistency calibration.}
        \label{fig:local_measurement}
    \end{subfigure}
    \caption{Evaluation of Consistency Calibration under Different Perturbation Settings.}
    \label{fig:toy_example}
    \vspace{-0.2in}
\end{figure}

Due to numerous types of data augmentations, determining the optimal perturbation strength using a continuous variable is challenging. To address this, we extend the perturbation process to the feature and logit levels by introducing noise with varying intensities. This approach yields effects similar to those observed with image-level perturbations, as demonstrated in Figure~\ref{fig:feature_disturb} and Figure~\ref{fig:logit_disturb}.

Interestingly, feature- and logit-level perturbations maintain significant calibration performance while offer huge computational advantages. With image-level perturbations, inference must be performed on the entire model \( T \) times. In contrast, feature-level require evaluating only the classification head, while logit-level only compute the argmax operation \( T \) times. This results in substantial reductions in computational costs. Experiments on other layers can be found in Appendix~\ref{Perturbation of different layer}.

\begin{proposition}
We propose a unified definition of our calibration methods, termed \textit{Consistency Calibration} (CC), which identifies perturbed neighbors at different levels. The calibrated prediction confidence score \(\hat{p}'\) is formally defined as:
\begin{equation}
\hat{p}'_k = \frac{1}{T} \sum_{t=1}^{T} \mathbbm{1}\left( \argmax q\left( \widetilde{h(x)}^t \right) = k \right), \quad \text{for} \, k = 1, \dots, K,
\label{eq:consistency_calibration}
\end{equation}
where \( h(x) \) is the representation of data \( x \), \( \widetilde{h(x)}^t \) is the perturbed representation, and \( q \) is the pipeline to extract the logits \( z \).
\end{proposition}
Specifically, for data-level perturbations: \( h(\cdot) = I(\cdot) \), \( \widetilde{h(x)} \) is the augmented data, \( q = g(f(\cdot)) \).
For feature-level perturbations: \( h(\cdot) = f(\cdot) \), \( \widetilde{h(x)}^t = h(x) + \epsilon_t \), \( q = g(\cdot) \).
For logit-level perturbations: \( h(\cdot) = g(f(\cdot)) \), \( \widetilde{h(x)}^t = h(x) + \epsilon_t \), \( q = I(\cdot) \). 
Here, \(I(\cdot)\) is the identity function, \( \epsilon_t \) represents the noise added to features or logits, with its strength determined by minimizing the ECE on a validation set. Given the strong calibration performance and computational efficiency of logit-level perturbations, we refer to logit-level consistency calibration as CC when no specification is provided.

\subsection{Consistency as a Local Uncertainty Estimation}
\label{Consistency as a Local Uncertainty Estimation}
Consistency-based methods do not rely on label information or additional data, as they generate their own neighborhood by perturbing the input data. This property allows consistency to serve as a criterion for instance-level uncertainty measurement. As illustrated in Figure~\ref{fig:local_measurement}, we examine a miscalibrated (incorrect prediction with high confidence) CIFAR-10 test sample, where a ResNet-50 model trained with Cross-Entropy (CE) shows overconfidence, assigning a confidence score of 0.997 despite being incorrect. Using optimal temperature, determined via a validation set, the confidence after temperature scaling decreases slightly, but the model remains overconfident at 0.903. 

For comparison, we apply CC by perturbing the logits (``CC (logits)"), applying train time data augmentation (``CC (Train Aug)"), and using a moderate augmentation method (``CC (Train Aug + Jitter)"). The confidence significantly decreases with these approaches. However, too strong augmentations may negatively impact model accuracy, which requires the need for a validation set to tune the augmentation strength, so we recommend using training-time augmentation to avoid the use of validation set while keeping the prediction accuracy.

Unlike many post-hoc calibration methods that require a large validation set to fine-tune hyperparameters, consistency-based confidence with train-time augmentation can directly provide calibrated confidence scores while maintaining recognizable by models. This approach is particularly valuable in data-limited scenarios, allowing consistency to produce an accurate local uncertainty estimation.

\subsection{Why consistency calibration works?}
\label{Why consistency calibration works?}
\begin{figure}[ht]
    \centering
    \begin{subfigure}[b]{0.325\textwidth}
        \centering
        \includegraphics[width=\textwidth]{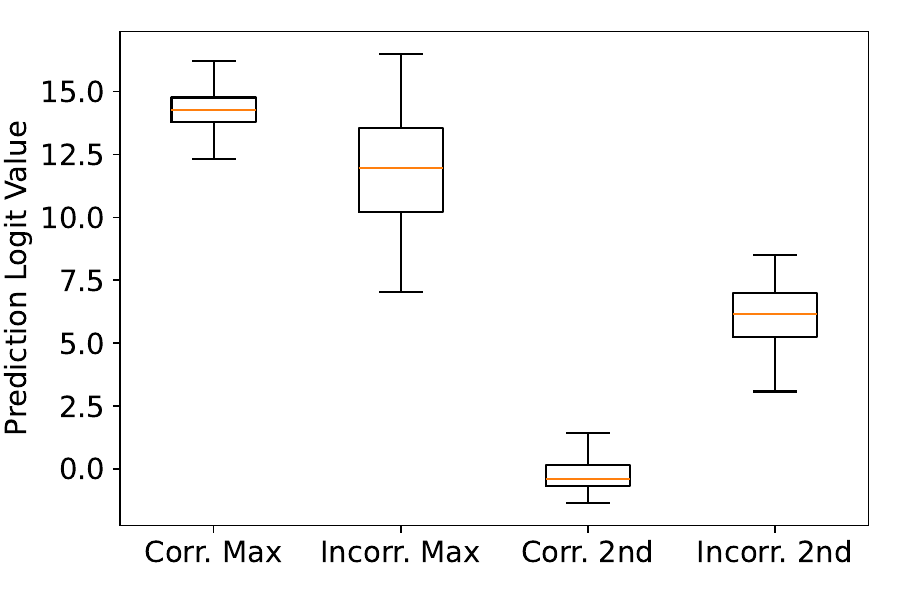}
        \caption{CIFAR-10}
        \label{fig:logit-boxplot-cifar10}
    \end{subfigure}
    \hfill
    \begin{subfigure}[b]{0.325\textwidth}
        \centering
        \includegraphics[width=\textwidth]{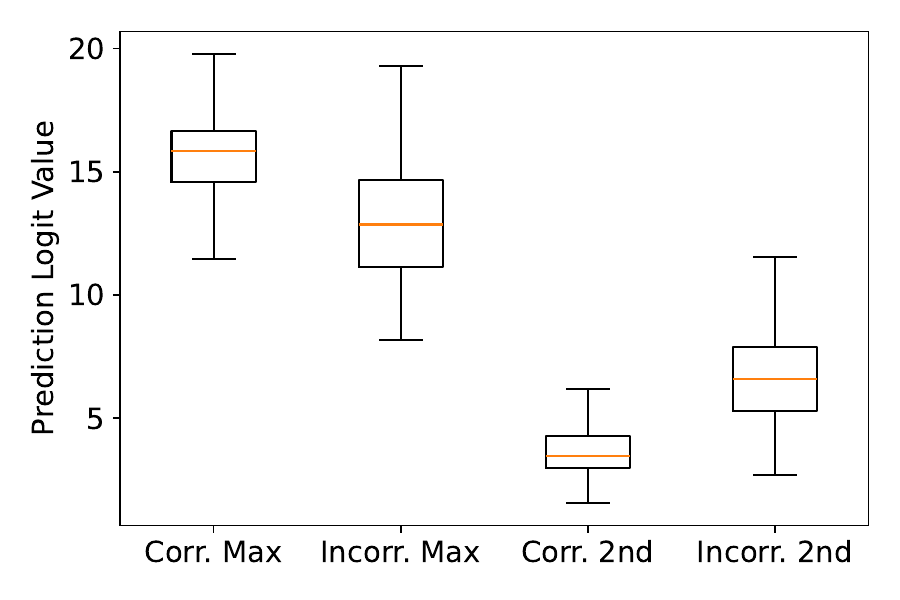}
        \caption{CIFAR-100}
        \label{fig:logit-boxplot-cifar100}
    \end{subfigure}
    \hfill
    \begin{subfigure}[b]{0.325\textwidth}
        \centering
        \includegraphics[width=\textwidth]{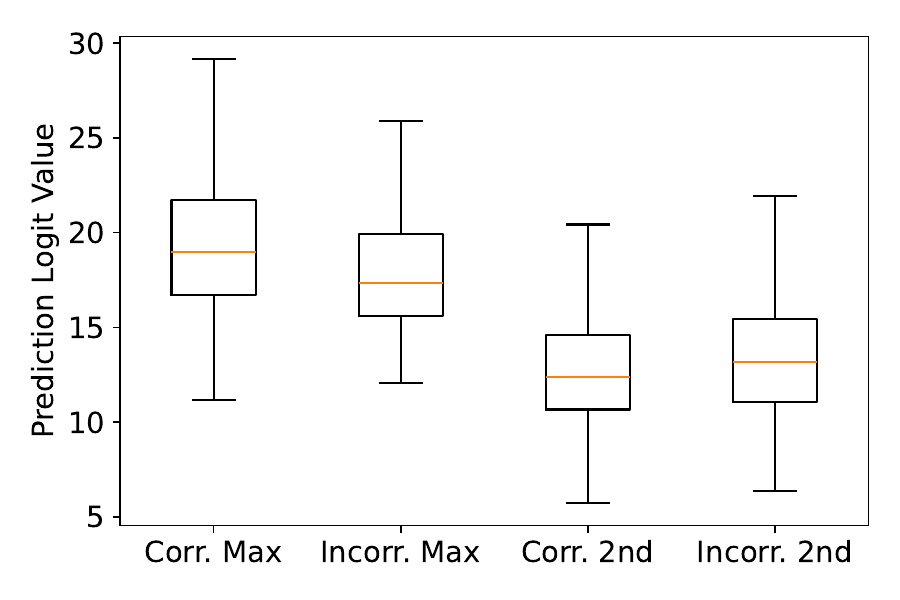}
        \caption{ImageNet-1K}
        \label{fig:logit-boxplot-imagenet}
    \end{subfigure}
    \caption{Distribution of the max logit and second-largest logit for correct and incorrect predictions with more than 99\% confidence, representing well-calibrated and miscalibrated samples on ResNet-50 across different datasets. The difference between the max logit and second-largest logit is significantly smaller for miscalibrated samples compared to well-calibrated samples.}
    \label{fig:logit-boxplot}
    \vspace{-0.2in}
\end{figure}

Perturbing images results in straightforward and intuitive image neighborhoods, but the effectiveness of perturbations at the logit level requires further explanation. To understand why logit perturbations work, we examined the differences between highly confident correct predictions and overconfident incorrect ones. These represent well-calibrated and poorly calibrated samples, respectively. During logit disturbance, the label with second-largest logit most likely to become the prediction label. To investigate this, we plotted box plots for both the maximum and second-largest logits for correct and incorrect predictions, as shown in Figure~\ref{fig:logit-boxplot-cifar10}.

For CIFAR-10 test samples, we selected predictions with confidence higher than 99\%. We refer to the maximum logit of correct predictions as ``Corr. Max" and that of incorrect predictions as ``Incorr. Max." Similarly, ``Corr. 2nd" represents the second-largest logit of correct predictions, while ``Incorr. 2nd" refers to the second-largest logit of incorrect predictions. As shown in Figure~\ref{fig:logit-boxplot-cifar10}, the maximum logit for correct predictions is significantly higher than for incorrect predictions. Additionally, the second-largest logit in correct predictions is much lower than that in incorrect predictions. This indicates that the gap between the maximum and second-largest logits is much larger for correct predictions than for incorrect ones. Despite large difference, due to softmax saturation, the model assigns abnormally high confidence (greater than 99\%) to both correct and incorrect predictions, leading to overconfident miscalibration.

Interestingly, we can leverage this difference in the logit gaps between correct and incorrect predictions. Perturbations can easily alter the predictions of overconfident, miscalibrated samples, while having minimal effect on well-calibrated, correct predictions. This different response to perturbations explains why consistency calibration is effective at the logit level. We observed similar patterns in experiments with CIFAR-100 and ImageNet, as shown in Figure~\ref{fig:logit-boxplot-cifar100} and Figure~\ref{fig:logit-boxplot-imagenet}.

\section{Experiments}
\subsection{Experimental Setup}

\paragraph{Datasets}  
We conduct experiments on several benchmark datasets, including CIFAR-10, CIFAR-100~\citep{krizhevsky2009learning}, and ImageNet~\citep{deng2009imagenet}. To assess calibration performance in data-imbalance scenarios, we also include ImageNet-LT~\citep{liu2019largescalelongtailedrecognitionopen}, characterized by its long-tailed class distribution. CIFAR-10 and CIFAR-100 contain 60,000 images of size $32 \times 32$ pixels, with 10 and 100 classes, respectively, split into 45,000 training, 5,000 for validation and 10,000 test images. For ImageNet-1K, we split 20\% of the original validation set as the new validation set, with the remainder used as the test set. We use the \(\epsilon\) searched on ImageNet-1k validation set to calibrate ImageNet-LT test set. The testing batch size for all datasets is set to 128.

\paragraph{Models}  
We evaluate our approach across various neural network architectures, including ResNet-50 and ResNet-110~\citep{he2016deep}, Wide ResNet~\citep{zagoruyko2016wide}, DenseNet-121~\citep{huang2017densely}, and Vision Transformers (ViT-B/16 and ViT-B/32)~\citep{dosovitskiy2021imageworth16x16words}. These models represent a diverse range of architectures and complexities, allowing us to assess the robustness of our method in different settings. For CIFAR-10 and CIFAR-100, we use pretrained weights from prior work~\citep{mukhoti2020calibrating}. All models are trained using stochastic gradient descent (SGD) with a momentum of 0.9 and weight decay of $5 \times 10^{-4}$ for 350 epochs.The learning rate is initialized at 0.1 for the first 150 epochs, reduced to 0.01 for the next 100, and further decreased to 0.001 for the final 100 epochs. For ImageNet, we use pretrained models from PyTorch~\citep{paszke2019pytorchimperativestylehighperformance}, following the training recipe available on PyTorch's model page.

\paragraph{Evaluation Metrics and Other Settings}  
Calibration performance is primarily evaluated using ECE, with additional metrics including AdaECE, CECE, Negative Log-Likelihood (NLL), and top-1 accuracy. All experiments are conducted on an NVIDIA 4090 GPU, with results averaged over five runs to ensure fairness. For all experiments, we set the number of perturbations to $T=1000$ and search the perturbation strength $\epsilon$ and noise type by minimizing ECE on the validation set.

\begin{table*}[ht]
\centering
\scriptsize
\begin{tabular*}{\textwidth}{@{\extracolsep{\fill}}ccccccccc}
\toprule
\textbf{Dataset} & \textbf{Model} & \textbf{Vanilla} & \textbf{TS} & \textbf{ETS} & \textbf{PTS} & \textbf{CTS} & \textbf{GC} & \textbf{CC (ours)} \\
\midrule
\multirow{2}{*}{CIFAR-10} 
 & ResNet-50      & 4.34 & 1.38 & 1.37 & 1.36 & 1.46 & 1.04 & \cellgray\textbf{0.78} \\ 
 % & ResNet-110     & 4.41 & 0.99 & 0.98 & \textbf{0.96} & 1.13 & 1.11 & \cellgray0.98\\ 
 % & DenseNet-121   & 4.51 & 1.41 & 1.41 & 1.38 & 1.44 & 1.28 & \cellgray\textbf{1.07}\\ 
 & Wide-ResNet    & 3.24 & 0.93 & 0.93 & 0.93 & 0.93 & 1.33 & \cellgray\textbf{0.36}\\ 
\midrule
\multirow{2}{*}{CIFAR-100} 
 & ResNet-50      & 17.52 & 5.71 & 5.68 & 5.64 & 6.05 & 3.55 & \cellgray\textbf{1.25}\\
 & Wide-ResNet    & 15.34 & 4.63 & 4.58 & 4.52 & 4.86 & 2.14 & \cellgray\textbf{1.61}\\ 
\midrule
\multirow{6}{*}{ImageNet-1K} 
 & ResNet-50     & 3.76 & 2.09 & 2.09 & 2.08 & 3.14 & 2.54 & \cellgray\textbf{1.53} \\ 
 & DenseNet-121   & 6.59 & 1.64 & 1.66 & 1.68 & 1.94 & 2.51 & \cellgray\textbf{1.48}\\ 
 & Wide-ResNet-50 & 5.49 & 3.03 & 3.04 & 3.04 & 4.13 & 2.16 & \cellgray\textbf{1.33}\\ 
 & Swin-B  & 5.02 & 3.90 & 3.90 & 3.93 & 5.43 & 1.61 & \cellgray\textbf{1.58} \\
 & ViT-B-16       & 5.61 & 3.61 & 3.62 & 3.64 & 5.50 & 1.75 & \cellgray\textbf{1.66}\\ 
 & ViT-B-32       & 6.40 & 3.76 & 3.78 & 3.84 & 5.74 & \textbf{1.39} & \cellgray1.72\\
\midrule
\multirow{6}{*}{ImageNet-LT}
 & ResNet-50     & 3.67 & 2.00 & 1.99 & 2.00 & 2.21 & 1.4  & \cellgray\textbf{1.24} \\ 
 & DenseNet-121  & 6.65 & 1.65 & 1.64 & 1.66 & 1.59 & 1.81 & \cellgray\textbf{1.23} \\ 
 & Wide-ResNet-50  & 5.39 & 2.97 & 2.96 & 2.96 & 3.52 & 1.49 & \cellgray\textbf{1.27} \\ 
 & Swin-B        & 4.66 & 4.02 & 4.03 & 4.08 & 5.02 & 1.66 & \cellgray\textbf{1.44} \\ 
 & ViT-B-16      & 5.57 & 3.61 & 3.62 & 3.64 & 4.94 & 1.76 & \cellgray\textbf{1.61} \\ 
 & ViT-B-32      & 5.15 & 5.67 & 5.67 & 5.68 & 5.71 & \textbf{1.48} & \cellgray{1.74} \\ 
\midrule
\bottomrule
\end{tabular*}
\caption{\textbf{Comparison of Post-Hoc Calibration Methods Using ECE\(\downarrow\) Across Various Datasets and Models.} ECE values are reported with 15 bins. The best-performing method for each dataset-model combination is in bold, and our method (CC) is highlighted. Results are averaged over 5 runs.}
\label{table:comparison_ECE}
\vspace{-0.2in}
\end{table*}

\subsection{Comparison with Post-Hoc Calibration Methods}  
We compare our proposed CC with widely used post-hoc calibration techniques, including Temperature Scaling (TS) \citep{guo2017calibration}, Ensemble Temperature Scaling (ETS) \citep{zhang2020mix}, Parameterized Temperature Scaling (PTS) \citep{tomani2022parameterized}, Class-based Temperature Scaling (CTS) \citep{frenkel2021network}, and Group Calibration (GC) \citep{yang2024beyond}, as well as uncalibrated models (Vanilla). Our evaluation covers CIFAR-10, CIFAR-100, ImageNet-1K, and ImageNet-LT, using various CNNs and transformers.

\paragraph{Calibration on Standard Datasets}  
CC consistently outperforms these methods across CIFAR-10, CIFAR-100, and ImageNet-1K, significantly reducing calibration error. The most notable improvement is seen in CIFAR-100, where CC excels while GC, despite its strong performance on other datasets, struggles. This highlights CC's robustness across datasets with varying complexities. CNNs, which often suffer from overconfidence, are generally well-calibrated with TS-based methods. However, transformers see limited calibration improvements from TS-based methods, with CC outperforming them by a large margin. On larger datasets like ImageNet-1K, CC maintains its advantage. Although GC slightly outperforms CC on ViT-B/32, it is computationally expensive due to the additional grouping process, whereas CC balances both efficiency and effectiveness.

\paragraph{Calibration on Long-Tail Datasets}  
On long-tail datasets like ImageNet-LT, TS-based models struggle to provide effective calibration, especially for transformers. For example, on ViT-B/32, TS-based methods fail to calibrate effectively, as they apply uniform adjustments across the dataset, smoothing or sharpening probabilities globally. In contrast, CC and GC perform well on long-tail datasets, particularly with transformers. GC excels due to its multicalibration~\citep{hebert2018multicalibration}, offering sample-wise adjustments, though it comes at a high computational cost. By leveraging local uncertainty estimation through input perturbations, CC better captures uncertainties in underrepresented tail classes, making it especially useful for handling imbalanced data scenarios.

\subsection{Calibration Performance on Other Metrics}
\begin{figure}[ht]
    \centering
    \includegraphics[width=\linewidth]{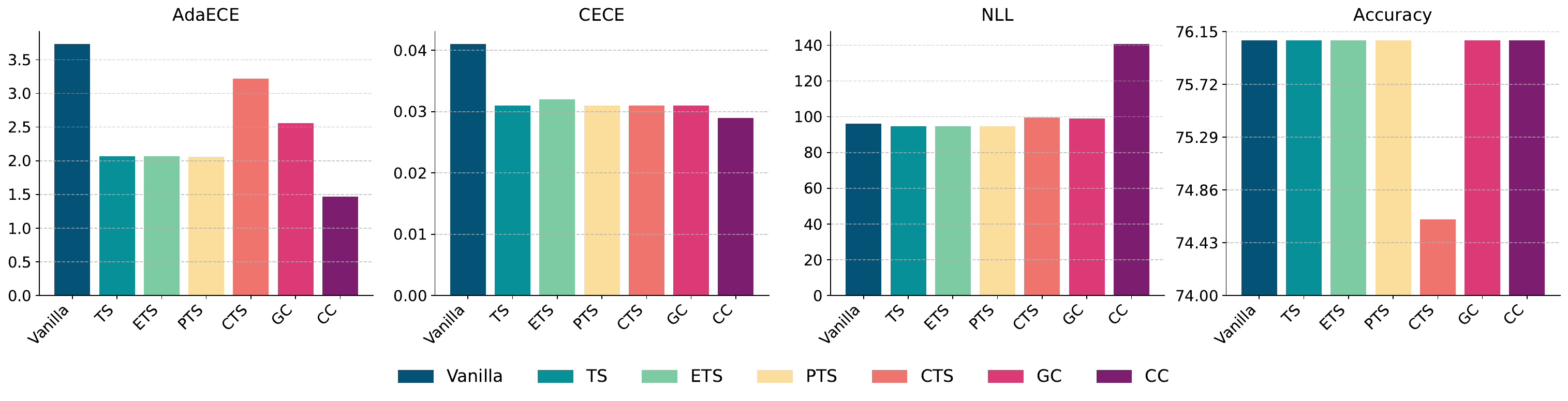}
    \caption{\textbf{Calibration performance of ResNet-50 on ImageNet-1K using AdaECE\(\downarrow\), CECE\(\downarrow\), NLL\(\downarrow\), and Accuracy\(\uparrow\).} ECE, AdaECE, and CECE are reported with 15 bins. Colors in the legend represent different methods. Results are averaged over 5 runs.}
    \label{fig:resnet50_imagenet}
    \vspace{-0.2in}
    
\end{figure}
We also evaluate CC using additional metrics: AdaECE, CECE, NLL, and accuracy to provide a comprehensive view of its performance. Results for ResNet-50 on ImageNet are shown here, with results for other models and datasets available in Appendix~\ref{appendix: Comparison of post-hoc calibration methods on other metrics}.
\vspace{-0.1in}
\paragraph{AdaECE and CECE}  
CC demonstrates superior performance on both AdaECE and CECE compared to traditional methods. AdaECE accounts for uneven confidence distributions, improving the reliability of ECE, while CECE gives detailed insights into classwise calibration. CC's strong results on both metrics show its effectiveness from different perspectives.
\vspace{-0.1in}
\paragraph{Accuracy Maintained}  
CC preserves the accuracy of the base models, showing no significant reduction in classification performance. As a post-hoc method, it does not require retraining, maintaining predictive capabilities, making it practical for real-world applications.
\vspace{-0.1in}
\paragraph{Increase in NLL}  
Interestingly, CC results in higher NLL values compared to other methods, reflecting a trade-off between calibration and the sharpness of probability estimates. This suggests that while CC reduces overconfidence in incorrect predictions, it also moderates overconfidence in correct predictions, leading to improved calibration without affecting accuracy.

\begin{table*}[ht]
\centering
\scriptsize
\begin{tabular*}{\textwidth}{@{\extracolsep{\fill}}cccccccccccccc}
\toprule
\textbf{Dataset} & \textbf{Model} &\multicolumn{2}{c}{\textbf{Cross-Entropy}} &\multicolumn{2}{c}{\textbf{Brier Loss}} &\multicolumn{2}{c}{\textbf{MMCE}} &\multicolumn{2}{c}{\textbf{LS-0.05}} &\multicolumn{2}{c}{\textbf{FLSD-53}} &\multicolumn{2}{c}{\textbf{FL-3}} \\
 & & base & ours & base & ours & base & ours & base & ours & base & ours & base & ours \\
\midrule
\multirow{4}{*}{CIFAR-10} & ResNet-50 & 4.34 & \cellgray \textbf{0.78} & 1.80 & \cellgray \textbf{1.07} & 4.56 & \cellgray \textbf{0.83} & 2.97 & \cellgray \textbf{1.24} & 1.55 & \cellgray \textbf{0.49} & 1.48 & \cellgray \textbf{0.66} \\ 
 & ResNet-110 & 4.41 & \cellgray \textbf{0.98} & 2.57 & \cellgray \textbf{0.48} & 5.08 & \cellgray \textbf{1.17} & \textbf{2.09} & \cellgray 2.30 & 1.88 & \cellgray \textbf{0.67} & 1.54 & \cellgray \textbf{0.48} \\ 
 & DenseNet-121 & 4.51 & \cellgray \textbf{1.07} & 1.52 & \cellgray \textbf{0.78} & 5.10 & \cellgray \textbf{1.18} & 1.87 & \cellgray \textbf{1.39} & 1.23 & \cellgray \textbf{0.68} & 1.31 & \cellgray \textbf{0.98} \\ 
 & Wide-ResNet & 3.24 & \cellgray \textbf{0.36} & 1.24 & \cellgray \textbf{0.58} & 3.29 & \cellgray \textbf{0.39} & 4.25 & \cellgray \textbf{1.15} & 1.58 & \cellgray \textbf{0.49} & 1.68 & \cellgray \textbf{0.53} \\ 
\midrule
\multirow{4}{*}{CIFAR-100} & ResNet-50 & 17.52 & \cellgray \textbf{1.25} & 6.57 & \cellgray \textbf{1.57} & 15.32 & \cellgray \textbf{1.98} & 7.82 & \cellgray \textbf{5.08} & 4.49 & \cellgray \textbf{1.43} & 5.16 & \cellgray \textbf{1.52} \\ 
 & ResNet-110 & 19.05 & \cellgray \textbf{4.57} & 7.88 & \cellgray \textbf{3.24} & 19.14 & \cellgray \textbf{4.41} & 11.04 & \cellgray \textbf{4.58} & 8.55 & \cellgray \textbf{3.47} & 8.64 & \cellgray \textbf{3.67} \\ 
 & DenseNet-121 & 20.99 & \cellgray \textbf{5.40} & 5.22 & \cellgray \textbf{1.82} & 19.10 & \cellgray \textbf{3.76} & 12.87 & \cellgray \textbf{4.99} & 3.70 & \cellgray \textbf{1.41} & 4.14 & \cellgray \textbf{1.94} \\ 
 & Wide-ResNet & 15.34 & \cellgray \textbf{1.61} & 4.34 & \cellgray \textbf{1.87} & 13.17 & \cellgray \textbf{2.17} & 4.89 & \cellgray \textbf{4.21} & 3.02 & \cellgray \textbf{1.64} & 2.14 & \cellgray \textbf{1.78} \\ 
\midrule
\bottomrule
\end{tabular*}
\caption{\textbf{Comparison of Train-time Calibration Methods Using ECE\(\downarrow\) Across Various Datasets and Models.} ECE values are reported with 15 bins. The best-performing method for each dataset-model combination is in bold, and our method (CC) is highlighted. Results are averaged over 5 runs.}
\label{table:CC_ece_compare_with_training_time_methods}
\vspace{-0.2in}
\end{table*}

\subsection{Comparison with Training-Time Calibration Methods}  
We evaluate CC alongside training-time calibration techniques, including Brier Loss~\citep{brier1950verification}, Maximum Mean Calibration Error (MMCE) \citep{kumar2018trainable}, Label Smoothing (LS-0.05)~\citep{szegedy2016rethinking}, and Focal Loss variants (FLSD-53 and FL-3)~\citep{mukhoti2020calibrating}, as shown in Table~\ref{table:CC_ece_compare_with_training_time_methods}. Our analysis shows that combining CC with these methods consistently enhances calibration performance across various models and datasets, further validating CC’s effectiveness alongside training-time approaches.

Moreover, as seen in Table~\ref{table:comparison_ECE}, CC alone, as a post-hoc calibration method, already outperforms these train-time techniques with minimal computational overhead, while train-time methods require significantly more resources. Additional results for other settings are available in Appendix~\ref{appendix: Comparison of various training-time calibration methods on other metrics}.

% \begin{table*}[ht]
% \centering
% \scriptsize
% \begin{tabular}{cccccccc}
% \toprule
%  \textbf{Model} & \textbf{Vanilla} & \textbf{TS} & \textbf{ETS} & \textbf{PTS} & \textbf{CTS} & \textbf{GC} & \textbf{CC} \\
% \midrule
%  ResNet-50  & 3.67 & 2.00 & 1.99 & 2.00 & 2.21 & 1.4 & \textbf{1.24} \\ 
%  DenseNet-121  & 6.65 & 1.65 & 1.64 & 1.66 & 1.59 & 1.81 & \textbf{1.23} \\ 
%  Wide-ResNet-50  & 5.39 & 2.97 & 2.96 & 2.96 & 3.52 & 1.49 & \textbf{1.27} \\ 
%  Swin-B  & 4.66 & 4.02 & 4.03 & 4.08 & 5.02 & 1.66 & \textbf{1.44} \\ 
%  ViT-B-16  & 5.57 & 3.61 & 3.62 & 3.64 & 4.94 & 1.76 & \textbf{1.61} \\ 
%  ViT-B-32  & 5.15 & 5.67 & 5.67 & 5.68 & 5.71 & \textbf{1.48} & 1.74 \\ 
% \midrule
% \bottomrule
% \end{tabular}
% \caption{Comparison of calibration methods using ECE on ImageNet-LT}
% \label{table:comparison_ECE_ImageNet-LT}
% \end{table*}
\subsection{Ablation Study}

\paragraph{Aggregation Methods}  
In our ablation study, we compare two aggregation methods for refining confidence estimates: the mean of softmax probabilities (\textit{Mean}), defined as:
\begin{equation}
\hat{p}_k = \frac{1}{T} \sum_{t=1}^{T} \softmax \left(q\left( \widetilde{h(x)}^t \right) \right), \quad \text{for} \, k = 1, \dots, K,
\end{equation}
and consistency-based aggregation (\textit{Consis.}) as shown in Eq.~\ref{eq:consistency_calibration}. Both methods leverage predictions over perturbed logits. The mean of softmax probabilities treats the perturbation process like an ensemble method, interpreting uncertainty as a distribution. We show the evaluation results on CIFAR-10 and CIFAR-100 in Table~\ref{table:ablation study - aggregation methods and noise type}. On smaller datasets like CIFAR-10, both methods perform similarly. However, on larger datasets with more classes, such as CIFAR-100 and ImageNet, consistency-based aggregation slightly outperforms softmax averaging. This suggests that consistency-based aggregation captures uncertainty better than the view of ensemble.

\paragraph{Choice of Noise}  
We investigate the impact of different noise types for input perturbations, comparing uniform noise (\textit{U}) and Gaussian noise (\textit{G}), as shown in Table~\ref{table:ablation study - aggregation methods and noise type}. Uniform noise performs better on datasets with fewer classes, such as CIFAR-10 and CIFAR-100. However, on larger datasets like ImageNet, Gaussian noise yields better results, likely due to variations in the gap between the maximum and second maximum logits across datasets as shown in Figure~\ref{fig:logit-boxplot}. The choice of noise is treated as a hyperparameter, offering flexibility to adapt to different datasets and models.

\paragraph{Number of Perturbations}  
We also assess the impact of the number of perturbations. As shown in Figure~\ref{fig:number_of_perturbations}, our experiments indicate that CC achieves strong calibration performance with as few as \(2^4 = 16\) perturbations. Although increasing the number of perturbations slightly improves results, the diminishing returns suggest that CC provides robust calibration with a moderate number of perturbations, ensuring both efficiency and accuracy.

\begin{table*}[ht]
\centering
\begin{minipage}{0.65\textwidth}
\centering
\scriptsize
\begin{tabular}{cccccc}
\toprule
\textbf{Dataset} & \textbf{Model} &\textbf{Mean U} & \textbf{Mean G} & \textbf{Consis. U} & \textbf{Consis. G} \\
\midrule
\multirow{2}{*}{CIFAR-10} 
 & ResNet-50  & \textbf{0.72} & 1.34& 0.78 & 1.33 \\ 
 & Wide-ResNet    & 0.37  & 0.80 & \textbf{0.36} & 0.83 \\ 
\midrule
\multirow{2}{*}{CIFAR-100} 
 & ResNet-50      & 1.52 & 2.70& \textbf{1.25} & 2.49\\ 
 & Wide-ResNet   & 1.86 & 2.08& \textbf{1.61} & 1.88  \\ 
\midrule
\multirow{2}{*}{ImageNet} 
 & ResNet-50   & 2.37&1.41&2.29&\textbf{1.27} \\ 
 & Wide-ResNet-50& 2.17&1.7&2.23&\textbf{1.57}\\ 
\midrule
\bottomrule
\end{tabular}
\caption{\textbf{Comparison of Aggregation Methods and Noise Types Using ECE\(\downarrow\) Across Various Datasets and Models.} ECE values are reported using 15 bins. The best-performing method for each dataset-model combination is highlighted in bold.}
\label{table:ablation study - aggregation methods and noise type}
\end{minipage}%
\hfill
\begin{minipage}{0.3\textwidth}
    \centering
    \includegraphics[width=\linewidth]{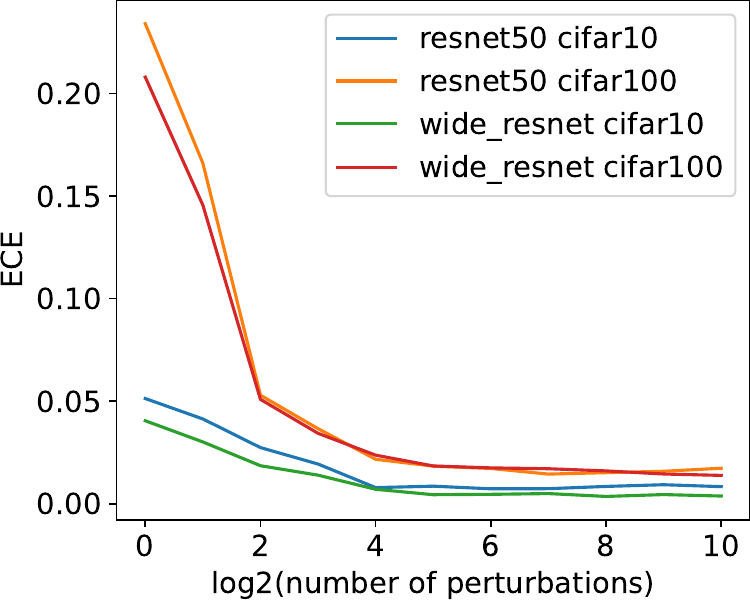}
    \caption{\textbf{Effect of Number of Perturbations on Calibration Performance (ECE\(\downarrow\))} }
    \label{fig:number_of_perturbations}
\end{minipage}
\vspace{-0.2in}
\end{table*}

% \begin{figure}
%     \centering
%     \includegraphics[width=0.32\linewidth]{figs/number_of_perturbations.pdf}
%     \caption{\textbf{Effect of Number of Perturbations on Calibration Performance (ECE\(\downarrow\))} The x-axis represents the number of perturbations (log2 scale), and the y-axis shows the ECE values. The results indicate that CC achieves strong calibration with as few as \(2^4 = 16\) perturbations.}
%     \label{fig:number_of_perturbations}
% \end{figure}
% \begin{table*}[ht]
% \centering
% \scriptsize
% \begin{tabular}{cccccc}
% \toprule
% \textbf{Dataset} & \textbf{Model} &\textbf{Mean U} & \textbf{Mean G} & \textbf{Consis. U} & \textbf{Consis. G} \\
% \midrule
% \multirow{2}{*}{CIFAR-10} 
%  & ResNet-50  & \textbf{0.72} & 1.34& 0.78 & 1.33 \\ 
%  & Wide-ResNet    & 0.37  & 0.80 & \textbf{0.36} & 0.83 \\ 
% \midrule
% \multirow{2}{*}{CIFAR-100} 
%  & ResNet-50      & 1.52 & 2.70& \textbf{1.25} & 2.49\\ 
%  & Wide-ResNet   & 1.86 & 2.08& \textbf{1.61} & 1.88  \\ 
% \midrule
% \multirow{2}{*}{ImageNet} 
%  & ResNet-50   & 2.37&1.41&2.29&\textbf{1.27} \\ 
%  & Wide-ResNet-50& 2.17&1.7&2.23&\textbf{1.57}\\ 
% \midrule
% \bottomrule
% \end{tabular}
% \caption{\textbf{Comparison of Aggregation Methods and Noise Types Using ECE\(\downarrow\) Across Various Datasets and Models.} ECE values are reported using 15 bins. The best-performing method for each dataset-model combination is highlighted in bold.}
% \label{table:ablation study - aggregation methods and noise type}
% \end{table*}

\section{Conclusion}
Consistency offers an alternative perspective on calibration by focusing on prediction stability under perturbations as an indicator of confidence. CC has proven highly effective in reducing calibration errors across various datasets. However, CC has limitations, such as the need for tuning perturbation strength and noise type, and its current focus on classification tasks, with its application to regression remaining unexplored. Future work can aim to develop a new, more universal consistency-based metric to complement existing metrics like ECE. This would provide a more comprehensive evaluation to calibration, ultimately leading to more reliable deep learning models.

\clearpage
\bibliography{iclr2025_conference}

\begin{thebibliography}{39}
\providecommand{\natexlab}[1]{#1}
\providecommand{\url}[1]{\texttt{#1}}
\expandafter\ifx\csname urlstyle\endcsname\relax
  \providecommand{\doi}[1]{doi: #1}\else
  \providecommand{\doi}{doi: \begingroup \urlstyle{rm}\Url}\fi

\bibitem[Brier(1950)]{brier1950verification}
Glenn~W Brier.
\newblock Verification of forecasts expressed in terms of probability.
\newblock \emph{Monthly weather review}, 78\penalty0 (1):\penalty0 1--3, 1950.

\bibitem[Chen et~al.(2018)Chen, Sahiner, Samuelson, Pezeshk, and Petrick]{chen2018calibration}
Weijie Chen, Berkman Sahiner, Frank Samuelson, Aria Pezeshk, and Nicholas Petrick.
\newblock Calibration of medical diagnostic classifier scores to the probability of disease.
\newblock \emph{Statistical methods in medical research}, 27\penalty0 (5):\penalty0 1394--1409, 2018.

\bibitem[Conde et~al.(2023)Conde, Barros, Lopes, Premebida, and Nunes]{conde2023approaching}
Pedro Conde, Tiago Barros, Rui~L Lopes, Cristiano Premebida, and Urbano~J Nunes.
\newblock Approaching test time augmentation in the context of uncertainty calibration for deep neural networks.
\newblock \emph{arXiv preprint arXiv:2304.05104}, 2023.

\bibitem[DeGroot \& Fienberg(1983)DeGroot and Fienberg]{degroot1983comparison}
Morris~H DeGroot and Stephen~E Fienberg.
\newblock The comparison and evaluation of forecasters.
\newblock \emph{Journal of the Royal Statistical Society: Series D (The Statistician)}, 32\penalty0 (1-2):\penalty0 12--22, 1983.

\bibitem[Deng et~al.(2009)Deng, Dong, Socher, Li, Li, and Fei-Fei]{deng2009imagenet}
Jia Deng, Wei Dong, Richard Socher, Li-Jia Li, Kai Li, and Li~Fei-Fei.
\newblock Imagenet: A large-scale hierarchical image database.
\newblock In \emph{2009 IEEE conference on computer vision and pattern recognition}, pp.\  248--255. Ieee, 2009.

\bibitem[Dosovitskiy et~al.(2021)Dosovitskiy, Beyer, Kolesnikov, Weissenborn, Zhai, Unterthiner, Dehghani, Minderer, Heigold, Gelly, Uszkoreit, and Houlsby]{dosovitskiy2021imageworth16x16words}
Alexey Dosovitskiy, Lucas Beyer, Alexander Kolesnikov, Dirk Weissenborn, Xiaohua Zhai, Thomas Unterthiner, Mostafa Dehghani, Matthias Minderer, Georg Heigold, Sylvain Gelly, Jakob Uszkoreit, and Neil Houlsby.
\newblock An image is worth 16x16 words: Transformers for image recognition at scale, 2021.
\newblock URL \url{https://arxiv.org/abs/2010.11929}.

\bibitem[Feng et~al.(2019)Feng, Rosenbaum, Glaeser, Timm, and Dietmayer]{feng2019can}
Di~Feng, Lars Rosenbaum, Claudius Glaeser, Fabian Timm, and Klaus Dietmayer.
\newblock Can we trust you? on calibration of a probabilistic object detector for autonomous driving.
\newblock \emph{arXiv preprint arXiv:1909.12358}, 2019.

\bibitem[Frenkel et~al.(2021)Frenkel, Goldberger, Goldberger, and Goldberger]{frenkel2021network}
Lior Frenkel, Jacob Goldberger, Jacob Goldberger, and Jacob Goldberger.
\newblock Network calibration by class-based temperature scaling.
\newblock In \emph{2021 29th European Signal Processing Conference (EUSIPCO)}, pp.\  1486--1490. IEEE, 2021.

\bibitem[Gal \& Ghahramani(2016)Gal and Ghahramani]{gal2016dropout}
Yarin Gal and Zoubin Ghahramani.
\newblock Dropout as a bayesian approximation: Representing model uncertainty in deep learning.
\newblock In \emph{international conference on machine learning}, pp.\  1050--1059. PMLR, 2016.

\bibitem[Geng et~al.(2023)Geng, Cai, Wang, Koeppl, Nakov, and Gurevych]{geng2023survey}
Jiahui Geng, Fengyu Cai, Yuxia Wang, Heinz Koeppl, Preslav Nakov, and Iryna Gurevych.
\newblock A survey of language model confidence estimation and calibration.
\newblock \emph{arXiv preprint arXiv:2311.08298}, 2023.

\bibitem[Guo et~al.(2017)Guo, Pleiss, Sun, and Weinberger]{guo2017calibration}
Chuan Guo, Geoff Pleiss, Yu~Sun, and Kilian~Q Weinberger.
\newblock On calibration of modern neural networks.
\newblock In \emph{International conference on machine learning}, pp.\  1321--1330. PMLR, 2017.

\bibitem[He et~al.(2016)He, Zhang, Ren, and Sun]{he2016deep}
Kaiming He, Xiangyu Zhang, Shaoqing Ren, and Jian Sun.
\newblock Deep residual learning for image recognition.
\newblock In \emph{Proceedings of the IEEE conference on computer vision and pattern recognition}, pp.\  770--778, 2016.

\bibitem[H{\'e}bert-Johnson et~al.(2018)H{\'e}bert-Johnson, Kim, Reingold, and Rothblum]{hebert2018multicalibration}
Ursula H{\'e}bert-Johnson, Michael Kim, Omer Reingold, and Guy Rothblum.
\newblock Multicalibration: Calibration for the (computationally-identifiable) masses.
\newblock In \emph{International Conference on Machine Learning}, pp.\  1939--1948. PMLR, 2018.

\bibitem[Huang et~al.(2017)Huang, Liu, Van Der~Maaten, and Weinberger]{huang2017densely}
Gao Huang, Zhuang Liu, Laurens Van Der~Maaten, and Kilian~Q Weinberger.
\newblock Densely connected convolutional networks.
\newblock In \emph{Proceedings of the IEEE conference on computer vision and pattern recognition}, pp.\  4700--4708, 2017.

\bibitem[Krizhevsky et~al.(2009)Krizhevsky, Hinton, et~al.]{krizhevsky2009learning}
Alex Krizhevsky, Geoffrey Hinton, et~al.
\newblock Learning multiple layers of features from tiny images.
\newblock \emph{arXiv preprint arXiv:2010.11929}, 2009.

\bibitem[Kull et~al.(2019)Kull, Silva~Filho, and Flach]{kull2019beyond}
Meelis Kull, Telmo Silva~Filho, and Peter Flach.
\newblock Beyond temperature scaling: Obtaining well-calibrated multiclass probabilities with dirichlet calibration.
\newblock \emph{Advances in neural information processing systems}, 32:\penalty0 12316--12326, 2019.

\bibitem[Kumar et~al.(2018)Kumar, Sarawagi, Jain, and Jain]{kumar2018trainable}
Aviral Kumar, Sunita Sarawagi, Ujjwal Jain, and Ujjwal Jain.
\newblock Trainable calibration measures for neural networks from kernel mean embeddings.
\newblock In \emph{International Conference on Machine Learning}, pp.\  2805--2814. PMLR, 2018.

\bibitem[Lakshminarayanan et~al.(2017)Lakshminarayanan, Pritzel, and Blundell]{lakshminarayanan2017simple}
Balaji Lakshminarayanan, Alexander Pritzel, and Charles Blundell.
\newblock Simple and scalable predictive uncertainty estimation using deep ensembles.
\newblock \emph{Advances in neural information processing systems}, 30, 2017.

\bibitem[Lin et~al.(2023)Lin, Trivedi, and Sun]{lin2023generating}
Zhen Lin, Shubhendu Trivedi, and Jimeng Sun.
\newblock Generating with confidence: Uncertainty quantification for black-box large language models.
\newblock \emph{arXiv preprint arXiv:2305.19187}, 2023.

\bibitem[Liu et~al.(2019)Liu, Miao, Zhan, Wang, Gong, and Yu]{liu2019largescalelongtailedrecognitionopen}
Ziwei Liu, Zhongqi Miao, Xiaohang Zhan, Jiayun Wang, Boqing Gong, and Stella~X. Yu.
\newblock Large-scale long-tailed recognition in an open world, 2019.
\newblock URL \url{https://arxiv.org/abs/1904.05160}.

\bibitem[Manakul et~al.(2023)Manakul, Liusie, and Gales]{manakul2023selfcheckgpt}
Potsawee Manakul, Adian Liusie, and Mark~JF Gales.
\newblock Selfcheckgpt: Zero-resource black-box hallucination detection for generative large language models.
\newblock \emph{arXiv preprint arXiv:2303.08896}, 2023.

\bibitem[Minderer et~al.(2021)Minderer, Djolonga, Romijnders, Hubis, Zhai, Houlsby, Tran, and Lucic]{minderer2021revisiting}
Matthias Minderer, Josip Djolonga, Rob Romijnders, Frances Hubis, Xiaohua Zhai, Neil Houlsby, Dustin Tran, and Mario Lucic.
\newblock Revisiting the calibration of modern neural networks.
\newblock \emph{Advances in Neural Information Processing Systems}, 34:\penalty0 15682--15694, 2021.

\bibitem[Mukhoti et~al.(2020)Mukhoti, Kulharia, Sanyal, Golodetz, Torr, and Dokania]{mukhoti2020calibrating}
Jishnu Mukhoti, Viveka Kulharia, Amartya Sanyal, Stuart Golodetz, Philip Torr, and Puneet Dokania.
\newblock Calibrating deep neural networks using focal loss.
\newblock \emph{Advances in Neural Information Processing Systems}, 33:\penalty0 15288--15299, 2020.

\bibitem[Naeini et~al.(2015)Naeini, Cooper, and Hauskrecht]{naeini2015obtaining}
Mahdi~Pakdaman Naeini, Gregory Cooper, and Milos Hauskrecht.
\newblock Obtaining well calibrated probabilities using bayesian binning.
\newblock In \emph{Proceedings of the AAAI conference on artificial intelligence}, volume~29, 2015.

\bibitem[Nixon et~al.(2019)Nixon, Dusenberry, Zhang, Jerfel, and Tran]{nixon2019measuring}
Jeremy Nixon, Michael~W Dusenberry, Linchuan Zhang, Ghassen Jerfel, and Dustin Tran.
\newblock Measuring calibration in deep learning.
\newblock In \emph{CVPR workshops}, volume~2, 2019.

\bibitem[Paszke et~al.(2019)Paszke, Gross, Massa, and Lerer]{paszke2019pytorchimperativestylehighperformance}
Adam Paszke, Sam Gross, Francisco Massa, and Adam Lerer.
\newblock Pytorch: An imperative style, high-performance deep learning library, 2019.
\newblock URL \url{https://arxiv.org/abs/1912.01703}.

\bibitem[Szegedy et~al.(2016)Szegedy, Vanhoucke, Ioffe, Shlens, and Wojna]{szegedy2016rethinking}
Christian Szegedy, Vincent Vanhoucke, Sergey Ioffe, Jon Shlens, and Zbigniew Wojna.
\newblock Rethinking the inception architecture for computer vision.
\newblock In \emph{Proceedings of the IEEE conference on computer vision and pattern recognition}, pp.\  2818--2826, 2016.

\bibitem[Tam et~al.(2022)Tam, Mascarenhas, Zhang, Kwan, Bansal, and Raffel]{tam2022evaluating}
Derek Tam, Anisha Mascarenhas, Shiyue Zhang, Sarah Kwan, Mohit Bansal, and Colin Raffel.
\newblock Evaluating the factual consistency of large language models through summarization.
\newblock \emph{arXiv preprint arXiv:2211.08412}, 2022.

\bibitem[Tao et~al.(2023{\natexlab{a}})Tao, Dong, Liu, Sun, and Xu]{tao2023calibrating}
Linwei Tao, Minjing Dong, Daochang Liu, Changming Sun, and Chang Xu.
\newblock Calibrating a deep neural network with its predecessors.
\newblock In \emph{Proceedings of the Thirty-Second International Joint Conference on Artificial Intelligence}, pp.\  4271--4279, 2023{\natexlab{a}}.

\bibitem[Tao et~al.(2023{\natexlab{b}})Tao, Dong, Xu, and Xu]{tao2023dual}
Linwei Tao, Minjing Dong, Chang Xu, and Chang Xu.
\newblock Dual focal loss for calibration.
\newblock In \emph{International Conference on Machine Learning}, pp.\  33833--33849. PMLR, 2023{\natexlab{b}}.

\bibitem[Tao et~al.(2023{\natexlab{c}})Tao, Zhu, Guo, Dong, and Xu]{tao2023benchmark}
Linwei Tao, Younan Zhu, Haolan Guo, Minjing Dong, and Chang Xu.
\newblock A benchmark study on calibration.
\newblock \emph{arXiv preprint arXiv:2308.11838}, 2023{\natexlab{c}}.

\bibitem[Tomani et~al.(2022)Tomani, Cremers, Buettner, and Sun]{tomani2022parameterized}
Christian Tomani, Daniel Cremers, Florian Buettner, and Yu~Sun.
\newblock Parameterized temperature scaling for boosting the expressive power in post-hoc uncertainty calibration.
\newblock In \emph{European Conference on Computer Vision}, pp.\  555--569. Springer, 2022.

\bibitem[Wang et~al.(2021)Wang, Feng, and Zhang]{wang2021rethinking}
Deng-Bao Wang, Lei Feng, and Min-Ling Zhang.
\newblock Rethinking calibration of deep neural networks: Do not be afraid of overconfidence.
\newblock \emph{Advances in Neural Information Processing Systems}, 34:\penalty0 11809--11820, 2021.

\bibitem[Wang et~al.(2022)Wang, Wei, Schuurmans, Le, Chi, Narang, Chowdhery, and Zhou]{wang2022self}
Xuezhi Wang, Jason Wei, Dale Schuurmans, Quoc Le, Ed~Chi, Sharan Narang, Aakanksha Chowdhery, and Denny Zhou.
\newblock Self-consistency improves chain of thought reasoning in language models.
\newblock \emph{arXiv preprint arXiv:2203.11171}, 2022.

\bibitem[Xiong et~al.(2023{\natexlab{a}})Xiong, Deng, Koh, Wu, Li, Xu, and Hooi]{xiong2023proximity}
Miao Xiong, Ailin Deng, Pang Wei~W Koh, Jiaying Wu, Shen Li, Jianqing Xu, and Bryan Hooi.
\newblock Proximity-informed calibration for deep neural networks.
\newblock \emph{Advances in Neural Information Processing Systems}, 36:\penalty0 68511--68538, 2023{\natexlab{a}}.

\bibitem[Xiong et~al.(2023{\natexlab{b}})Xiong, Hu, Lu, Li, Fu, He, and Hooi]{xiong2023can}
Miao Xiong, Zhiyuan Hu, Xinyang Lu, Yifei Li, Jie Fu, Junxian He, and Bryan Hooi.
\newblock Can llms express their uncertainty? an empirical evaluation of confidence elicitation in llms.
\newblock \emph{arXiv preprint arXiv:2306.13063}, 2023{\natexlab{b}}.

\bibitem[Yang et~al.(2024)Yang, Zhan, Gan, and Sun]{yang2024beyond}
Jia-Qi Yang, De-Chuan Zhan, Le~Gan, and Yu~Sun.
\newblock Beyond probability partitions: Calibrating neural networks with semantic aware grouping.
\newblock \emph{Advances in Neural Information Processing Systems}, 36, 2024.

\bibitem[Zagoruyko \& Komodakis(2016)Zagoruyko and Komodakis]{zagoruyko2016wide}
Sergey Zagoruyko and Nikos Komodakis.
\newblock Wide residual networks.
\newblock \emph{arXiv preprint arXiv:1605.07146}, 2016.

\bibitem[Zhang et~al.(2020)Zhang, Kapishnikov, Singh, and Poon]{zhang2020mix}
Hong Zhang, Alexey Kapishnikov, Amar Singh, and Honglak Poon.
\newblock Mix-n-match: Ensemble and compositional methods for uncertainty calibration in deep learning.
\newblock \emph{arXiv preprint arXiv:2010.08092}, 2020.

\end{thebibliography}
\bibliographystyle{iclr2025_conference}

\clearpage

\appendix
\section{Related Works}
Numerous studies have explored the phenomenon of overconfidence in modern neural networks and investigated their calibration properties ~\citep{guo2017calibration, minderer2021revisiting, wang2021rethinking, tao2023benchmark}. Calibration methods can generally be categorized into two main approaches: post-hoc methods and train-time calibration methods.

\paragraph{Calibration Methods} Post-hoc calibration methods adjust model outputs after training to improve calibration. A widely used technique is Temperature Scaling (TS) \citep{guo2017calibration}, which smooths softmax probabilities by search a temperature factor on a validation set. Enhanced variants of TS include Parameterized Temperature Scaling (PTS) \citep{tomani2022parameterized}, which uses a neural network to learn the temperature, and Class-based Temperature Scaling (CTS) \citep{frenkel2021network}, which applies adjustments on a class-wise basis. Group Calibration (GC) \citep{yang2024beyond} and ProCal \citep{xiong2023proximity} aim for multi-calibration \citep{hebert2018multicalibration} by splitting data samples by proximity and grouping. Another stream of work is train-time calibration such as Brier Loss~\citep{brier1950verification}, Dirichlet Scaling \citep{kull2019beyond},  Maximum Mean Calibration Error (MMCE) \citep{kumar2018trainable}, Label Smoothing~\citep{szegedy2016rethinking}, and Focal Loss~\citep{mukhoti2020calibrating} and Dual Focal Loss~\citep{tao2023dual}. ~\citet{tao2023calibrating} propose to use a new training framework to improve calibration. However, these methods often require substantial higher computational overhead.

\paragraph{Ensemble-Based Calibration} Ensemble-based methods ensemble multiple outputs in different ways. They use models or samples to approximate Bayesian Inference. \citet{lakshminarayanan2017simple} propose deep ensembles as a scalable alternative to Bayesian Neural Networks (BNNs) for uncertainty estimation. Similarly, \citet{gal2016dropout} treat dropout as approximate Bayesian inference. Data-centric ensemble techniques using test-time augmentation, as described by \citet{conde2023approaching}, also help improve calibration. \citet{zhang2020mix} resort to the power of Bayesian inference and proposed a Ensemble-based TS (ETS). However, these methods typically require significant computational resources to train multiple models or perform repeated inferences. In contrast, our approach relies on consistency rather than probability distribution modeling.

\paragraph{Consistency in LLMs} Consistency has emerged as a key approach for black-box uncertainty estimation and hallucination detection in large language models (LLMs). These methods evaluate uncertainty by measuring variability in outputs across slight changes, such as different sampling techniques or rephrased prompts. Confident models produce stable outputs, while variability indicates uncertainty. For instance, SelfCheckGPT \citep{manakul2023selfcheckgpt} uses sampling and similarity metrics like BERTScore and NLI to detect hallucinations, while \citet{lin2023generating} analyze a similarity matrix to estimate uncertainty. \citet{xiong2023can} further break down uncertainty estimation into prompting, sampling, and consistency-based aggregation. These methods, which rely on output stability, are efficient alternatives to probabilistic approaches.

\section{Perturbation of different layer}
\label{Perturbation of different layer}

This section presents a detailed analysis of the impact of perturbations applied at various levels of a ResNet50 model, trained on CIFAR-10. The experiments were conducted using 32 samples, and the effects on ECE, accuracy, and optimal perturbation values were evaluated.

\begin{table*}[ht]
\centering
\scriptsize
\begin{tabular}{cccc}
\toprule
\textbf{Perturbation Level} & \textbf{ECE (\%)} & \textbf{Accuracy (\%)} & \textbf{Optimal Perturbation} \\
\midrule
Image & 1.1& 95.25 & train aug jitter0.1 \\ 
Logits & 0.73 & 95.04 & 8.2 \\ 
Feature (Last Layer) & 2.06& 95.06 & 3.0 \\ 
Feature (Layer 4) & 0.53& 95.29 & 13.28 \\ 
Feature (Layer 3) & 53.12& 10.03 & 20.12 \\ 
Feature (Layer 2) & 56.28& 10.02 & 20.21 \\ 
Feature (Layer 1) & 49.53& 10.11 & 20.75 \\ 
\midrule
\bottomrule
\end{tabular}
\caption{Comparison of perturbations at different layers with number of samples set to 32 using ECE\(\downarrow\) and Accuracy\(\uparrow\), evaluated on ResNet50 with CIFAR-10. ECE values are reported with 15 bins. Optimal Perturbations for logits and features are represented in $\epsilon$ value}
\label{table:ablation study}
\end{table*}

From Table \ref{table:ablation study}, we observe a clear trend in the performance of perturbations applied at different layers of the model. Perturbation at the logits level achieves a favorable trade-off between calibration and efficiency. Although the perturbation applied to the fourth layer's feature space slightly improves the ECE to 0.53\%, the associated computational cost is significantly higher, with the optimal perturbation value of 13.28. 

On the other hand, perturbations applied at lower feature levels (Layer 1 to Layer 3) result in severe degradation of both accuracy and calibration. Specifically, the ECE increases drastically to above 50\%, and accuracy drops to approximately 10\%, with a significant increase in computing time and memory use. This suggests that perturbing the features at these lower layers disrupts the model's ability to recognize patterns and correctly classify the input data. We hypothesize that this is due to the higher sensitivity of lower layers to the raw data structure, where perturbations may significantly distort the features necessary for effective recognition.

% 1. table header formal

% 2. layer4 works, but cost too much, we use logits to balance efficiency

% 3. other layer not working, we guess the perturbation make the model can not recognise the data at lower layer

% 4. caption: number of samples set to 32, on resnet50 cifar10

\section{Comparison of post-hoc calibration methods on other metrics}
\label{appendix: Comparison of post-hoc calibration methods on other metrics}

As shown in table \ref{table:comparison_ADAECE}, The proposed CC method consistently achieves the lowest AdaECE values, outperforming the other methods. This indicates better calibration performance, in line with our discussion in the main text. For instance, in CIFAR-10, Wide-ResNet has an AdaECE of 0.40 with CC compared to 3.24 for Vanilla, showing a significant improvement. Similar results are observed across other models and datasets. The formula for Adaptive-ECE is as follows:

\begin{equation}
\text{Adaptive-ECE}=\sum_{i=1}^{B}\frac{|B_i|}{N} \left| I_i - C_i \right| \text{ s.t. } \forall i,j\cdot|B_i|=|B_j|
\end{equation}

\begin{table}[ht]
\centering
\scriptsize
\begin{tabular*}{\textwidth}{@{\extracolsep{\fill}}ccccccccc}
\toprule
 \textbf{Dataset} & \textbf{Model} & \textbf{Vanilla} & \textbf{TS} & \textbf{ETS} & \textbf{PTS} & \textbf{CTS} & \textbf{GC} & \textbf{CC (ours)}\\

\midrule
\multirow{4}{*}{CIFAR-10} & ResNet-50 & 4.33 & 2.14 & 2.14 & 2.14 & 1.71 & 1.24 & \cellgray\textbf{0.64} \\ 
 & ResNet-110 & 4.40 & 1.89 & 1.89 & 1.90 & 1.31 & \textbf{0.94} & \cellgray0.96 \\ 
 & DenseNet-121 & 4.49 & 2.12 & 2.12 & 2.12 & 1.71 & 1.28 & \cellgray\textbf{1.20} \\ 
 & Wide-ResNet & 3.24 & 1.71 & 1.71 & 1.71 & 1.42 & 1.17 & \cellgray\textbf{0.40} \\ 
\midrule
\multirow{2}{*}{CIFAR-100} & ResNet-50 & 17.52 & 5.76 & 5.72 & 5.66 & 5.79 & 3.43 & \cellgray\textbf{1.61} \\ 
 & Wide-ResNet & 15.34 & 4.48 & 4.45 & 4.41 & 4.69 & 2.24 & \cellgray\textbf{1.73} \\ 
\midrule
\multirow{5}{*}{ImageNet} & ResNet-50  & 3.73 & 2.07 & 2.07 & 2.06 & 3.22 & 2.56 & \cellgray\textbf{1.47} \\ 
 & DenseNet-121  & 6.59 & 1.67 & 1.68 & 1.69 & 1.89 & 2.49 & \cellgray\textbf{1.36} \\ 
 & Wide-ResNet-50  & 5.32 & 2.97 & 2.97 & 2.95 & 4.13 & 2.18 & \cellgray\textbf{1.27} \\ 
 & ViT-B-16  & 5.59 & 4.05 & 4.06 & 4.08 & 5.50 & 1.86 & \cellgray\textbf{1.76} \\ 
 & ViT-B-32  & 6.40 & 3.83 & 3.85 & 3.91 & 5.73 & \textbf{1.33} & \cellgray1.77 \\ 
\midrule
\bottomrule
\end{tabular*}
\caption{\textbf{Comparison of Post-Hoc Calibration Methods Using AdaECE\(\downarrow\) Across Various Datasets and Models.} AdaECE values are reported with 15 bins. The best results for each  combination is in bold, and our method (CC) is highlighted. Results are averaged over 5 runs.}
\label{table:comparison_ADAECE}
\end{table}

As shown in table \ref{table:comparison_CECE}, The CC method also performs the best in terms of class-wise calibration, with consistently lower CECE values. This confirms that CC provides better calibration across individual classes, as discussed in the main body. For example, for ResNet-50 on CIFAR-100, CC achieves a CECE of 0.20, which is the lowest among the methods.
CECE is another measure of calibration performance that addresses the deficiency of ECE in only measuring the calibration performance of the single predicted class.
It can be formulated as:
\begin{equation}
\text{Classwise-ECE}=\frac{1}{\mathcal{K}} \sum_{i=1}^{B} \sum_{j=1}^{\mathcal{K}} \frac{|B_{i,j}|}{N} \left|I_{i,j} - C_{i,j} \right|
\end{equation}

\begin{table}[ht]
\centering
\scriptsize
\begin{tabular*}{\textwidth}{@{\extracolsep{\fill}}ccccccccc}
\toprule
 \textbf{Dataset} & \textbf{Model} & \textbf{Vanilla} & \textbf{TS} & \textbf{ETS} & \textbf{PTS} & \textbf{CTS} & \textbf{GC} & \textbf{CC (ours)}\\

\midrule
\multirow{4}{*}{CIFAR-10} & ResNet-50 & 0.91 & 0.45 & 0.45 & 0.45 & 0.41 & 0.46 & \cellgray\textbf{0.39} \\ 
 & ResNet-110 & 0.92 & 0.48 & 0.48 & 0.48 & 0.42 & 0.52 & \cellgray\textbf{0.41} \\ 
 & DenseNet-121 & 0.92 & 0.48 & 0.48 & 0.48 & \textbf{0.41} & 0.54 & \cellgray0.43 \\ 
 & Wide-ResNet & 0.68 & 0.37 & 0.37 & 0.37 & 0.37 & 0.48 & \cellgray\textbf{0.32} \\ 
\midrule
\multirow{2}{*}{CIFAR-100} & ResNet-50 & 0.38 & 0.21 & 0.21 & 0.21 & 0.22 & 0.21 & \cellgray\textbf{0.20} \\ 
 & Wide-ResNet & 0.34 & 0.19 & 0.19 & 0.19 & 0.20 & 0.20 & \cellgray\textbf{0.18} \\ 
\midrule
\multirow{5}{*}{ImageNet} & ResNet-50  & \textbf{0.03} & \textbf{0.03} & \textbf{0.03} & \textbf{0.03} & \textbf{0.03} & \textbf{0.03} & \cellgray\textbf{0.03} \\ 
 & DenseNet-121  & \textbf{0.03} & \textbf{0.03} & \textbf{0.03} & \textbf{0.03} & \textbf{0.03} & \textbf{0.03} & \cellgray\textbf{0.03} \\ 
 & Wide-ResNet-50  & 0.03 & 0.03 & 0.03 & 0.03 & 0.03 & 0.03 & \cellgray\textbf{0.02} \\ 
 & ViT-B-16  & 0.03 & \textbf{0.02} & \textbf{0.02} & \textbf{0.02} & 0.03 & 0.02 & \cellgray\textbf{0.02} \\ 
 & ViT-B-32  & \textbf{0.03} & \textbf{0.03} & \textbf{0.03} & \textbf{0.03} & \textbf{0.03} & \textbf{0.03} & \cellgray\textbf{0.03} \\ 
\midrule
\bottomrule
\end{tabular*}
\caption{\textbf{Comparison of Post-Hoc Calibration Methods Using CECE\(\downarrow\) Across Various Datasets and Models.} CECE values are reported with 15 bins. The best-performing method for each dataset-model combination is in bold, and our method (CC) is highlighted. Results are averaged over 5 runs.}
\label{table:comparison_CECE}
\end{table}

As shown in table \ref{table:comparison_NLL}, interestingly, the NLL values are generally higher with the CC method compared to some other calibration methods, despite its superior calibration performance in AdaECE and CECE. This suggests that while CC improves calibration, it may come at the cost of slightly higher NLL values. For instance, for CIFAR-100 on ResNet-50, CC has a higher NLL than TS, but it remains competitive overall.
\begin{table}[ht]
\centering
\scriptsize
\begin{tabular*}{\textwidth}{@{\extracolsep{\fill}}ccccccccc}
\toprule
 \textbf{Dataset} & \textbf{Model} & \textbf{Vanilla} & \textbf{TS} & \textbf{ETS} & \textbf{PTS} & \textbf{CTS} & \textbf{GC} & \textbf{CC (ours)}\\

\midrule
\multirow{4}{*}{CIFAR-10} & ResNet-50 & 41.21 & 20.39 & 20.39 & 20.38 & 20.15 & \textbf{19.97} & \cellgray20.39 \\ 
 & ResNet-110 & 47.52 & 21.52 & 21.52 & 21.52 & 20.84 & \textbf{20.68} & \cellgray23.33 \\ 
 & DenseNet-121 & 42.93 & 21.78 & 21.78 & 21.78 & 21.01 & \textbf{20.30} & \cellgray22.19 \\ 
 & Wide-ResNet & 26.75 & 15.33 & 15.33 & 15.33 & \textbf{15.13} & 15.32 & \cellgray17.10 \\ 
\midrule
\multirow{2}{*}{CIFAR-100} & ResNet-50 & 153.67 & \textbf{106.07} & \textbf{106.07} & \textbf{106.07} & 106.25 & 107.80 & \cellgray108.40 \\ 
 & Wide-ResNet & 140.11 & \textbf{95.71} & \textbf{95.71} & \textbf{95.71} & 96.38 & 96.92 & \cellgray99.30 \\ 
\midrule
\multirow{5}{*}{ImageNet} & ResNet-50  & 96.12 & 94.82 & 94.82 & \textbf{94.81} & 99.58 & 99.07 & \cellgray140.57 \\ 
 & DenseNet-121  & 109.52 & \textbf{103.90} & \textbf{103.90} & 103.91 & 106.13 & 108.14 & \cellgray162.02 \\ 
 & Wide-ResNet-50  & 88.56 & \textbf{86.46} & \textbf{86.46} & \textbf{86.46} & 91.68 & nan & \cellgray120.59 \\ 
 & ViT-B-16  & 83.71 & 78.63 & 78.63 & \textbf{78.63} & 85.19 & 82.14 & \cellgray106.89 \\ 
 & ViT-B-32  & 107.76 & 101.67 & 101.67 & \textbf{101.66} & 107.53 & 105.45 & \cellgray141.71 \\ 
\midrule
\bottomrule
\end{tabular*}
\caption{\textbf{Comparison of Post-Hoc Calibration Methods Using NLL\(\downarrow\) Across Various Datasets and Models.}  The best-performing method for each dataset-model combination is in bold, and our method (CC) is highlighted. Results are averaged over 5 runs.}
\label{table:comparison_NLL}
\end{table}

\ref{table:comparison_ACCURACY} indicates that there is little to no change in accuracy across the calibration methods, with all methods performing similarly in terms of classification accuracy. This patter is consistent with the main section, showing CC improves calibration without sacrificing accuracy. For example, on CIFAR-10, Wide-ResNet achieves almost identical accuracy for all methods, with CC slightly outperforming others in specific cases.

\begin{table}[ht]
\centering
\scriptsize
\begin{tabular*}{\textwidth}{@{\extracolsep{\fill}}ccccccccc}
\toprule
 \textbf{Dataset} & \textbf{Model} & \textbf{Vanilla} & \textbf{TS} & \textbf{ETS} & \textbf{PTS} & \textbf{CTS} & \textbf{GC} & \textbf{CC (ours)}\\

\midrule
\multirow{4}{*}{CIFAR-10} & ResNet-50 & 95.05 & 95.05 & 95.05 & 95.05 & 94.98 & 95.05 & \cellgray\textbf{95.06} \\ 
 & ResNet-110 & 95.11 & 95.11 & 95.11 & 95.11 & \textbf{95.18} & 95.11 & \cellgray95.16 \\ 
 & DenseNet-121 & 95.02 & 95.02 & 95.02 & 95.02 & 95.01 & 95.02 & \cellgray\textbf{95.04} \\ 
 & Wide-ResNet & \textbf{96.13} & \textbf{96.13} & \textbf{96.13} & \textbf{96.13} & 96.06 & \textbf{96.13} & \cellgray\textbf{96.13} \\ 
\midrule
\multirow{2}{*}{CIFAR-100} & ResNet-50 & 76.70 & 76.70 & 76.70 & 76.70 & \textbf{76.72} & 76.70 & \cellgray76.71 \\ 
 & Wide-ResNet & 79.29 & 79.29 & 79.29 & 79.29 & 79.17 & 79.29 & \cellgray\textbf{79.31} \\ 
\midrule
\multirow{5}{*}{ImageNet} & ResNet-50  & \textbf{76.08} & \textbf{76.08} & \textbf{76.08} & \textbf{76.08} & 74.62 & \textbf{76.08} & \cellgray\textbf{76.08} \\ 
 & DenseNet-121  & 74.16 & 74.16 & 74.16 & 74.16 & 73.08 & 74.16 & \cellgray\textbf{74.37} \\ 
 & Wide-ResNet-50  & 78.40 & 78.40 & 78.40 & 78.40 & 77.07 & 78.40 & \cellgray\textbf{78.48} \\ 
 & ViT-B-16  & \textbf{81.09} & \textbf{81.09} & \textbf{81.09} & \textbf{81.09} & 80.01 & \textbf{81.09} & \cellgray81.06 \\ 
 & ViT-B-32  & \textbf{75.94} & \textbf{75.94} & \textbf{75.94} & \textbf{75.94} & 74.90 & \textbf{75.94} & \cellgray75.90 \\ 
\midrule
\bottomrule
\end{tabular*}
\caption{\textbf{Comparison of Post-Hoc Calibration Methods Using Accuracy\(\uparrow\) Across Various Datasets and Models.} Top-1 accuracy values are reported. The best results for each  combination is in bold, and our method (CC) is highlighted. Results are averaged over 5 runs.}
\label{table:comparison_ACCURACY}
\end{table}

\begin{figure}[ht]
    \centering
    \includegraphics[width=\linewidth]{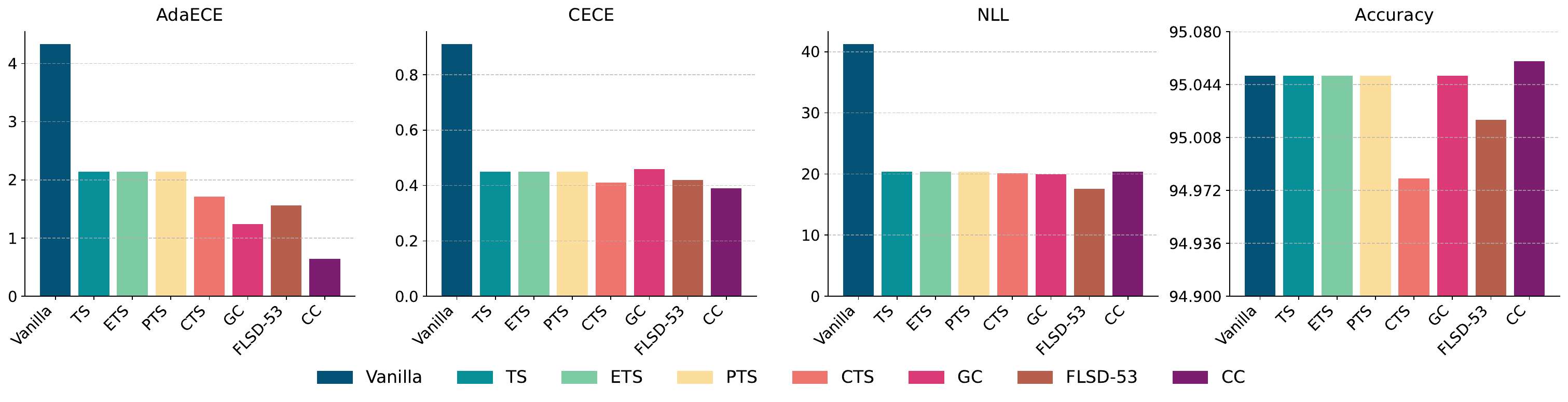}
    \caption{\textbf{Calibration performance of ResNet-50 on Cifar-10 using AdaECE\(\downarrow\), CECE\(\downarrow\), NLL\(\downarrow\), and Accuracy\(\uparrow\).} ECE, AdaECE, and CECE are reported with 15 bins. Colors in the legend represent different methods. Results are averaged over 5 runs.}
    \label{fig:resnet50_cifar10}
\end{figure}

In figure \ref{fig:resnet50_cifar10}, we see that the proposed CC method significantly reduces both AdaECE and CECE values compared to other calibration methods, indicating better calibration for Wide-ResNet on CIFAR-10. The accuracy remains mostly unchanged across all methods, while NLL is slightly higher for CC compared to other methods like TS and ETS. This behavior is consistent with our findings in the main text.

In Figure \ref{fig:resnet50_cifar10}, for ResNet-50 on CIFAR-10, the CC method demonstrates excellent performance with the lowest AdaECE and CECE values, further supporting its effectiveness in calibration. NLL is higher for CC, which is interesting given its superior performance in other metrics. However, accuracy remains largely unchanged, consistent with the overall findings discussed in the text.

\begin{figure}[ht]
    \centering
    \includegraphics[width=\linewidth]{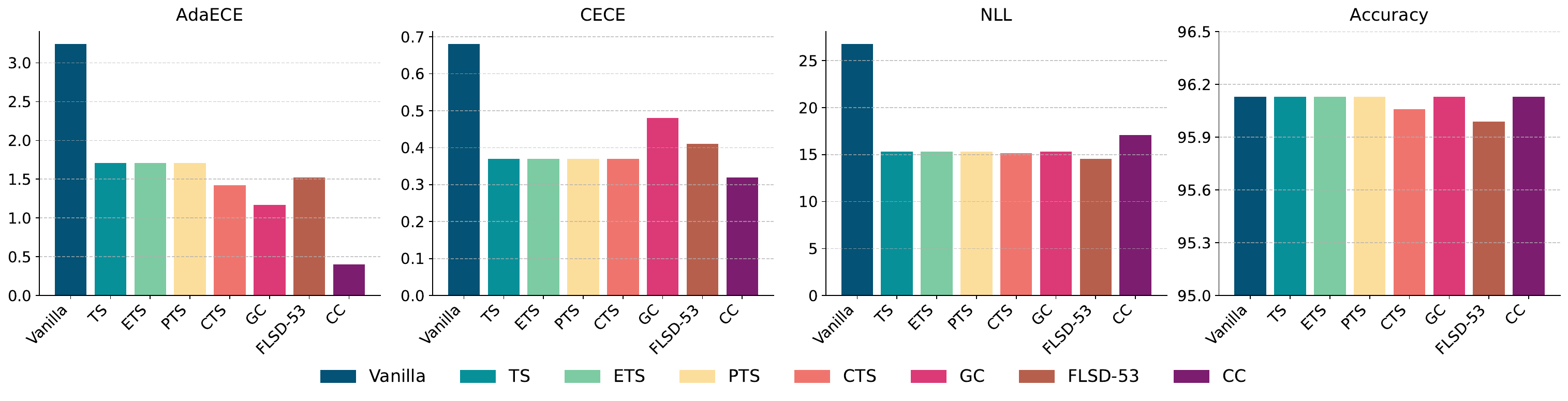}
    \caption{\textbf{Calibration performance of Wide-ResNet on CIFAR-10 using AdaECE\(\downarrow\), CECE\(\downarrow\), NLL\(\downarrow\), and Accuracy\(\uparrow\).} ECE, AdaECE, and CECE are reported with 15 bins. Colors in the legend represent different methods. Results are averaged over 5 runs.}
    \label{fig:wideresnet_cifar10}
\end{figure}
\begin{figure}[ht]
    \centering
    \includegraphics[width=\linewidth]{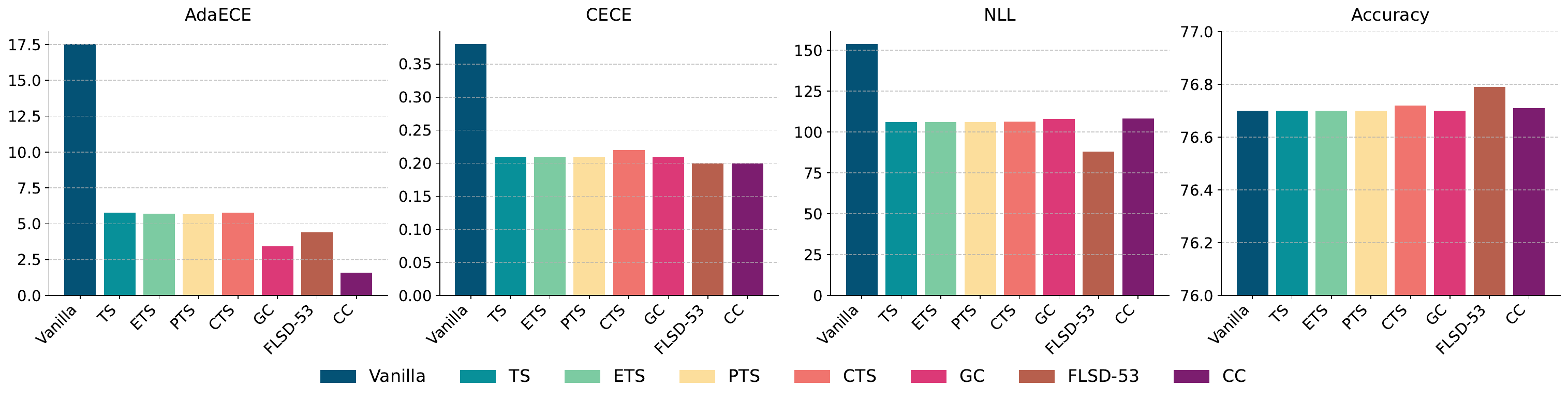}
    \caption{\textbf{Calibration performance of ResNet-50 on CIFAR-100 using AdaECE\(\downarrow\), CECE\(\downarrow\), NLL\(\downarrow\), and Accuracy\(\uparrow\).} ECE, AdaECE, and CECE are reported with 15 bins. Colors in the legend represent different methods. Results are averaged over 5 runs.}
    \label{fig:resnet50_cifar100}
\end{figure}

Figure \ref{fig:resnet50_cifar100} illustrates the performance of ResNet-50 on CIFAR-100 across different calibration methods. The proposed CC method again shows the lowest AdaECE and CECE, confirming its superior calibration performance. NLL for CC is slightly higher compared to TS, but accuracy shows minimal changes across methods. These results align with our overall conclusions that CC improves calibration without sacrificing accuracy.

\section{Comparison of various training-time calibration methods on other metrics}
\label{appendix: Comparison of various training-time calibration methods on other metrics}
As shown in Table~\ref{table:CC_adaece_compare_with_training_time_methods}, CC consistently outperforms baseline models across all metrics and datasets. Specifically, on CIFAR-10 and CIFAR-100, CC achieves significantly lower AdaECE scores for ResNet-50, ResNet-110, DenseNet-121, and Wide-ResNet compared to traditional methods such as Brier Loss, and MMCE. For instance, on CIFAR-100 with ResNet-110, CC reduces the AdaECE from 19.05 (baseline) to 5.28, showing superior calibration performance.

\begin{table*}[ht]
\centering
\scriptsize
\begin{tabular*}{\textwidth}{@{\extracolsep{\fill}}cccccccccccccc}
\toprule
\textbf{Dataset} & \textbf{Model} &\multicolumn{2}{c}{\textbf{Cross-Entropy}} &\multicolumn{2}{c}{\textbf{Brier Loss}} &\multicolumn{2}{c}{\textbf{MMCE}} &\multicolumn{2}{c}{\textbf{LS-0.05}} &\multicolumn{2}{c}{\textbf{FLSD-53}} &\multicolumn{2}{c}{\textbf{FL-3}} \\
 & & base & ours & base & ours & base & ours & base & ours & base & ours & base & ours \\
\midrule
\multirow{4}{*}{CIFAR-10} & ResNet-50 & 4.33 & \cellgray \textbf{0.64} & 1.75 & \cellgray \textbf{0.99} & 4.55 & \cellgray \textbf{1.06} & 3.88 & \cellgray \textbf{1.74} & 1.56 & \cellgray \textbf{0.36} & 1.95 & \cellgray \textbf{0.71} \\ 
 & ResNet-110 & 4.40 & \cellgray \textbf{0.96} & 2.60 & \cellgray \textbf{0.30} & 5.07 & \cellgray \textbf{1.80} & 4.48 & \cellgray \textbf{2.43} & 2.08 & \cellgray \textbf{0.73} & 1.64 & \cellgray \textbf{0.38} \\ 
 & DenseNet-121 & 4.49 & \cellgray \textbf{1.20} & 2.02 & \cellgray \textbf{0.64} & 5.10 & \cellgray \textbf{1.76} & 4.40 & \cellgray \textbf{1.94} & 1.38 & \cellgray \textbf{0.53} & 1.23 & \cellgray \textbf{0.69} \\ 
 & Wide-ResNet & 3.24 & \cellgray \textbf{0.40} & 1.70 & \cellgray \textbf{0.57} & 3.29 & \cellgray \textbf{0.63} & 4.27 & \cellgray \textbf{1.54} & 1.52 & \cellgray \textbf{0.42} & 1.84 & \cellgray \textbf{0.42} \\ 
\midrule
\multirow{4}{*}{CIFAR-100} & ResNet-50 & 17.52 & \cellgray \textbf{1.61} & 6.55 & \cellgray \textbf{1.90} & 15.32 & \cellgray \textbf{1.88} & 7.66 & \cellgray \textbf{6.17} & 4.39 & \cellgray \textbf{1.48} & 5.09 & \cellgray \textbf{1.70} \\ 
 & ResNet-110 & 19.05 & \cellgray \textbf{5.28} & 7.72 & \cellgray \textbf{3.54} & 19.14 & \cellgray \textbf{5.14} & 11.14 & \cellgray \textbf{8.00} & 8.56 & \cellgray \textbf{3.50} & 8.64 & \cellgray \textbf{3.98} \\ 
 & DenseNet-121 & 20.99 & \cellgray \textbf{5.85} & 5.04 & \cellgray \textbf{2.02} & 19.10 & \cellgray \textbf{3.90} & 12.83 & \cellgray \textbf{7.06} & 3.54 & \cellgray \textbf{1.52} & 4.14 & \cellgray \textbf{2.03} \\ 
 & Wide-ResNet & 15.34 & \cellgray \textbf{1.73} & 4.28 & \cellgray \textbf{1.92} & 13.16 & \cellgray \textbf{2.06} & 5.14 & \cellgray \textbf{4.75} & 2.77 & \cellgray \textbf{1.79} & 2.07 & \cellgray \textbf{1.58} \\ 
\midrule
\bottomrule
\end{tabular*}
\caption{\textbf{Comparison of Train-time Calibration Methods Using AdaECE\(\downarrow\) Across Various Datasets and Models.} AdaECE values are reported with 15 bins. The best results for each  combination is in bold, and our method (CC) is highlighted. Results are averaged over 5 runs.}
\label{table:CC_adaece_compare_with_training_time_methods}
\end{table*}

In Table~\ref{table:CC_cece_compare_with_training_time_methods}, the CECE results further reinforce the effectiveness of CC across all metrics. For CIFAR-10, CC improves CECE for all models compared to baseline methods. For instance, with ResNet-50, the CECE decreases from 0.91 to 0.39. Similar trends are observed on CIFAR-100, with Wide-ResNet showing a reduction in CECE from 0.34 (baseline) to 0.18 when using CC, demonstrating enhanced class-wise calibration.

\begin{table*}[ht]
\centering
\scriptsize
\begin{tabular*}{\textwidth}{@{\extracolsep{\fill}}cccccccccccccc}
\toprule
\textbf{Dataset} & \textbf{Model} &\multicolumn{2}{c}{\textbf{Cross-Entropy}} &\multicolumn{2}{c}{\textbf{Brier Loss}} &\multicolumn{2}{c}{\textbf{MMCE}} &\multicolumn{2}{c}{\textbf{LS-0.05}} &\multicolumn{2}{c}{\textbf{FLSD-53}} &\multicolumn{2}{c}{\textbf{FL-3}} \\
 & & base & ours & base & ours & base & ours & base & ours & base & ours & base & ours \\
\midrule
\multirow{4}{*}{CIFAR-10} & ResNet-50 & 0.91 & \cellgray \textbf{0.39} & 0.46 & \cellgray \textbf{0.35} & 0.94 & \cellgray \textbf{0.47} & 0.71 & \cellgray \textbf{0.53} & 0.42 & \cellgray \textbf{0.35} & 0.43 & \cellgray \textbf{0.39} \\ 
 & ResNet-110 & 0.92 & \cellgray \textbf{0.41} & 0.59 & \cellgray \textbf{0.41} & 1.04 & \cellgray \textbf{0.50} & \textbf{0.66} & \cellgray 0.67 & 0.48 & \cellgray \textbf{0.39} & 0.43 & \cellgray \textbf{0.37} \\ 
 & DenseNet-121 & 0.92 & \cellgray \textbf{0.43} & 0.46 & \cellgray \textbf{0.37} & 1.04 & \cellgray \textbf{0.59} & 0.60 & \cellgray \textbf{0.48} & 0.41 & \cellgray \textbf{0.35} & 0.42 & \cellgray \textbf{0.35} \\ 
 & Wide-ResNet & 0.68 & \cellgray \textbf{0.32} & 0.44 & \cellgray \textbf{0.32} & 0.70 & \cellgray \textbf{0.38} & 0.79 & \cellgray \textbf{0.41} & 0.41 & \cellgray \textbf{0.28} & 0.44 & \cellgray \textbf{0.30} \\ 
\midrule
\multirow{4}{*}{CIFAR-100} & ResNet-50 & 0.38 & \cellgray \textbf{0.20} & 0.22 & \cellgray \textbf{0.19} & 0.34 & \cellgray \textbf{0.18} & 0.23 & \cellgray \textbf{0.22} & 0.20 & \cellgray \textbf{0.19} & 0.20 & \cellgray \textbf{0.19} \\ 
 & ResNet-110 & 0.41 & \cellgray \textbf{0.21} & 0.24 & \cellgray \textbf{0.19} & 0.42 & \cellgray \textbf{0.20} & 0.26 & \cellgray \textbf{0.22} & 0.24 & \cellgray \textbf{0.19} & 0.24 & \cellgray \textbf{0.20} \\ 
 & DenseNet-121 & 0.45 & \cellgray \textbf{0.23} & 0.20 & \cellgray 0.20 & 0.42 & \cellgray \textbf{0.23} & 0.29 & \cellgray \textbf{0.22} & 0.19 & \cellgray 0.19 & 0.20 & \cellgray \textbf{0.19} \\ 
 & Wide-ResNet & 0.34 & \cellgray \textbf{0.18} & 0.19 & \cellgray \textbf{0.18} & 0.30 & \cellgray \textbf{0.17} & 0.21 & \cellgray \textbf{0.19} & 0.18 & \cellgray \textbf{0.17} & 0.18 & \cellgray \textbf{0.17} \\ 
\midrule
\bottomrule
\end{tabular*}
\caption{\textbf{Comparison of Train-time Calibration Methods Using CECE\(\downarrow\) Across Various Datasets and Models.} CECE values are reported with 15 bins. The best results for each  combination is in bold, and our method (CC) is highlighted. Results are averaged over 5 runs.}
\label{table:CC_cece_compare_with_training_time_methods}
\end{table*}

Table~\ref{table:CC_nll_compare_with_training_time_methods} presents the NLL comparison. It is interesting as mentioned in the main section, the CC method sometimes produces higher NLL values. 

\begin{table*}[ht]
\centering
\scriptsize
% Reduce inter-column spacing
\setlength{\tabcolsep}{5pt}
\begin{tabular*}{\textwidth}{@{}l l c c c c c c c c c c c c@{}}
\toprule
\textbf{Dataset} & \textbf{Model} &
\multicolumn{2}{c}{\textbf{Cross-Entropy}} &
\multicolumn{2}{c}{\textbf{Brier Loss}} &
\multicolumn{2}{c}{\textbf{MMCE}} &
\multicolumn{2}{c}{\textbf{LS-0.05}} &
\multicolumn{2}{c}{\textbf{FLSD-53}} &
\multicolumn{2}{c}{\textbf{FL-3}} \\
 & & \textbf{Base} & \textbf{Ours} &
 \textbf{Base} & \textbf{Ours} &
 \textbf{Base} & \textbf{Ours} &
 \textbf{Base} & \textbf{Ours} &
 \textbf{Base} & \textbf{Ours} &
 \textbf{Base} & \textbf{Ours} \\
\midrule
\multirow{4}{*}{CIFAR-10} 
 & ResNet-50 & 41.2 & \cellgray \textbf{20.4} & \textbf{18.7} & \cellgray 22.3 & 44.8 & \cellgray \textbf{20.9} & \textbf{27.7} & \cellgray 29.3 & \textbf{17.6} & \cellgray 22.7 & \textbf{18.4} & \cellgray 24.2 \\ 
 & ResNet-110 & 47.5 & \cellgray \textbf{25.5} & \textbf{20.4} & \cellgray 22.5 & 55.7 & \cellgray \textbf{25.5} & 29.9 & \cellgray \textbf{29.4} & \textbf{18.5} & \cellgray 21.9 & \textbf{17.8} & \cellgray 23.1 \\ 
 & DenseNet-121 & 42.9 & \cellgray \textbf{24.0} & \textbf{19.1} & \cellgray 21.2 & 52.1 & \cellgray \textbf{31.2} & 28.7 & \cellgray \textbf{28.5} & \textbf{18.4} & \cellgray 27.2 & \textbf{18.0} & \cellgray 28.3 \\ 
 & Wide-ResNet & 26.8 & \cellgray \textbf{17.1} & \textbf{15.9} & \cellgray 16.2 & 28.5 & \cellgray \textbf{18.2} & \textbf{21.7} & \cellgray 24.5 & \textbf{14.6} & \cellgray 17.6 & \textbf{15.2} & \cellgray 19.9 \\ 
\midrule
\multirow{4}{*}{CIFAR-100} 
 & ResNet-50 & 153.7 & \cellgray \textbf{113.0} & \textbf{99.6} & \cellgray 133.5 & 125.3 & \cellgray \textbf{116.7} & \textbf{121.0} & \cellgray 133.9 & \textbf{88.0} & \cellgray 128.8 & \textbf{87.5} & \cellgray 128.1 \\ 
 & ResNet-110 & 179.2 & \cellgray \textbf{122.3} & \textbf{110.7} & \cellgray 146.9 & 180.6 & \cellgray \textbf{125.3} & \textbf{133.1} & \cellgray 141.4 & \textbf{89.9} & \cellgray 126.9 & \textbf{90.9} & \cellgray 132.0 \\ 
 & DenseNet-121 & 205.6 & \cellgray \textbf{163.1} & \textbf{98.3} & \cellgray 139.9 & 166.6 & \cellgray \textbf{146.8} & \textbf{142.0} & \cellgray 185.8 & \textbf{85.5} & \cellgray 129.0 & \textbf{87.1} & \cellgray 130.8 \\ 
 & Wide-ResNet & 140.1 & \cellgray \textbf{102.5} & \textbf{84.6} & \cellgray 98.7 & 119.6 & \cellgray \textbf{109.3} & \textbf{108.1} & \cellgray 136.6 & \textbf{76.9} & \cellgray 108.7 & \textbf{74.7} & \cellgray 106.8 \\ 
\midrule
\bottomrule
\end{tabular*}
\caption{\textbf{Comparison of Train-time Calibration Methods Using NLL\(\downarrow\) Across Various Datasets and Models.} The best-performing method for each dataset-model combination is in bold, and our method (CC) is highlighted. Results are averaged over 5 runs.}
\label{table:CC_nll_compare_with_training_time_methods}
\end{table*}

Table~\ref{table:CC_accuracy_compare_with_training_time_methods} presents a comparison of classification accuracies. While achieving superior calibration performance by CC, the accuracy remains unaffected across all metrics.

\begin{table*}[ht]
\centering
\scriptsize
\setlength{\tabcolsep}{5pt}
\begin{tabular*}{\textwidth}{@{\extracolsep{\fill}}cccccccccccccc}
\toprule
\textbf{Dataset} & \textbf{Model} &\multicolumn{2}{c}{\textbf{Cross-Entropy}} &\multicolumn{2}{c}{\textbf{Brier Loss}} &\multicolumn{2}{c}{\textbf{MMCE}} &\multicolumn{2}{c}{\textbf{LS-0.05}} &\multicolumn{2}{c}{\textbf{FLSD-53}} &\multicolumn{2}{c}{\textbf{FL-3}} \\
 & & base & ours & base & ours & base & ours & base & ours & base & ours & base & ours \\
\midrule
\multirow{4}{*}{CIFAR-10} & ResNet-50 & \textbf{95.05} & \cellgray 95.06 & \textbf{94.99} & \cellgray 95.01 & 95.01 & \cellgray \textbf{94.99} & 94.71 & \cellgray \textbf{94.68} & 95.02 & \cellgray \textbf{94.95} & 94.75 & \cellgray 94.75 \\ 
 & ResNet-110 & \textbf{95.11} & \cellgray 95.16 & 94.52 & \cellgray \textbf{94.48} & \textbf{94.60} & \cellgray 94.63 & \textbf{94.48} & \cellgray 94.49 & \textbf{94.57} & \cellgray 94.63 & \textbf{94.92} & \cellgray 94.94 \\ 
 & DenseNet-121 & 95.02 & \cellgray \textbf{95.01} & 94.90 & \cellgray \textbf{94.86} & \textbf{94.59} & \cellgray 94.60 & 94.91 & \cellgray 94.91 & 94.58 & \cellgray \textbf{94.51} & 94.66 & \cellgray 94.66 \\ 
 & Wide-ResNet & 96.13 & \cellgray \textbf{96.12} & 95.92 & \cellgray \textbf{95.90} & 96.09 & \cellgray \textbf{96.05} & \textbf{95.80} & \cellgray 95.83 & \textbf{95.99} & \cellgray 96.01 & 95.87 & \cellgray 95.87 \\ 
\midrule
\multirow{4}{*}{CIFAR-100} & ResNet-50 & \textbf{76.70} & \cellgray 76.71 & 76.60 & \cellgray \textbf{76.58} & 76.80 & \cellgray 76.80 & \textbf{76.56} & \cellgray 76.65 & 76.79 & \cellgray \textbf{76.73} & \textbf{77.24} & \cellgray 77.34 \\ 
 & ResNet-110 & 77.27 & \cellgray \textbf{77.17} & 74.91 & \cellgray \textbf{74.79} & \textbf{76.93} & \cellgray 76.96 & \textbf{76.57} & \cellgray 76.64 & \textbf{77.48} & \cellgray 77.49 & 77.08 & \cellgray \textbf{77.04} \\ 
 & DenseNet-121 & \textbf{75.47} & \cellgray 75.49 & \textbf{76.27} & \cellgray 76.30 & 76.03 & \cellgray 76.03 & \textbf{75.94} & \cellgray 75.96 & 77.34 & \cellgray 77.34 & \textbf{76.76} & \cellgray 76.85 \\ 
 & Wide-ResNet & 79.29 & \cellgray \textbf{79.25} & 79.43 & \cellgray \textbf{79.29} & 79.27 & \cellgray \textbf{79.23} & \textbf{78.83} & \cellgray 78.88 & \textbf{79.91} & \cellgray 79.92 & \textbf{80.30} & \cellgray 80.34 \\ 
\midrule
\bottomrule
\end{tabular*}
\caption{\textbf{Comparison of Train-time Calibration Methods Using Accuracy\(\uparrow\) Across Various Datasets and Models.} Top-1 Accuracy values are reported. The best results for each  combination is in bold, and our method (CC) is highlighted. Results are averaged over 5 runs.}
\label{table:CC_accuracy_compare_with_training_time_methods}
\end{table*}

\end{document}